\documentclass[journal,twoside,web]{ieeecolor}
\usepackage{generic}
\usepackage{cite}

\usepackage{microtype}
\let\labelindent\relax
\usepackage{amsmath,amssymb,amsfonts, enumitem}
\usepackage{bbm, dsfont}
\usepackage{graphicx}
\usepackage{textcomp}
\usepackage{mathtools}
\usepackage{algorithm, algpseudocode}
\usepackage{subcaption}
\usepackage{wrapfig}

\usepackage{booktabs}
\usepackage{tabularx}
\usepackage[T1]{fontenc}
\usepackage[utf8]{inputenc}
\newcolumntype{Y}{>{\centering\arraybackslash}X}

\newtheorem{lemma}{Lemma}
\newtheorem{theorem}{Theorem}

\newtheorem{assumption}{Assumption}
\newtheorem{remark}{Remark}

\newtheorem{proposition}{Proposition}

\newcommand{\bt}[1]{{\boldsymbol{b}_t^{\scalebox{.6}{$(#1)$}}}}
\newcommand{\ttat}[1]{\boldsymbol{\theta}_t^{\scalebox{.6}{$(#1)$}}}

\DeclareMathOperator*{\argmin}{argmin}

\newcommand{\bbE}{\mathbb{E}}
\newcommand{\bbR}{\mathbb{R}}
\newcommand{\indic}{\mathds{1}}
\newcommand{\Tgd}{T_{\mathrm{GD}}}

\newcommand{\degg}{\mathrm{deg}_g}
\newcommand{\nn}{\nonumber}

\newcommand{\ccon}{N^{\mathrm{Con}}}
\newcommand{\cucon}{N^{\mathrm{Ucon}}} 
\newcommand{\cpcon}{N^{\mathrm{Con}}_{\mathrm{proj}}}
\newcommand{\cpucon}{N^{\mathrm{Ucon}}_{\mathrm{proj}}} 
\newcommand{\Ccon}{D^{\mathrm{Con}}}
\newcommand{\Cucon}{D^{\mathrm{Ucon}}} 
\newcommand{\Cpcon}{D^{\mathrm{Con}}_{\mathrm{proj}}}
\newcommand{\Cpucon}{D^{\mathrm{Ucon}}_{\mathrm{proj}}} 
\newcommand{\svdeq}{\overset{\mathrm{SVD}}=} %{\stackrel{EVD}{=}}
\newcommand{\qreq}{\overset{\mathrm{QR}}=} %{\stackrel{EVD}{=}}
\newcommand{\scalemath}[2]{\scalebox{#1}{\mbox{\ensuremath{\displaystyle #2}}}}
\newcommand{\SD}[2]{\mathrm{SD}_2(#1,#2)}
\newcommand{\AvgCons}{\textsc{Agree}_g}
\newcommand{\gradU}[1]{\mathrm{gradU}^{\scalebox{.6}{$(#1)$}}}

\newcommand{\Err}[1]{\mathrm{Err}^{\scalebox{.6}{$(#1)$}}}
\newcommand{\Errg}[1]{\mathrm{Err}_{g}^{\scalebox{.6}{$(#1)$}}}
\newcommand{\Errgp}[1]{\mathrm{Err}_{g'}^{\scalebox{.6}{$(#1)$}}} 
\newcommand{\ConsErr}[2]{\mathrm{ConsErr}_{#1}^{\scalebox{.6}{$(#2)$}}}
\newcommand{\UconsErr}[2]{\mathrm{UconsErr}_{\scalebox{.6}{$#1$}}^{\scalebox{.6}{$(#2)$}}}
\newcommand{\InpErr}[2]{\mathrm{InpErr}_{#1}^{\scalebox{.6}{$(#2)$}}}
\newcommand{\econ}{\epsilon_{\mathrm{con}}}
\newcommand{\Tcon}{T_{\mathrm{con}}}
\newcommand{\Tpm}{T_{\mathrm{pm}}}
\newcommand{\Tcongd}{T_{\mathrm{con,GD}}}
\newcommand{\Tconinit}{T_{\mathrm{con,init}}}
\newcommand{\ninit}{n_{\mathrm{init}}}
\newcommand{\ngd}{n_{\mathrm{GD}}}
\newcommand{\Utilde}[2]{\boldsymbol{\widetilde{U}}_{#1}^{\scalebox{.6}{$(#2)$}}}
\newcommand{\Ubreve}[2]{\boldsymbol{\breve{U}}_{#1}^{\scalebox{.6}{$(#2)$}}}
\newcommand{\Uhat}[1]{\boldsymbol{\widehat{U}}^{\scalebox{.6}{$(#1)$}} }
\newcommand{\Ubar}[1]{\boldsymbol{\bar{U}}^{\scalebox{.6}{$(#1)$}} }
\newcommand{\Ustar}{{\boldsymbol{U}^\star}}
\newcommand{\Utrue}{{\boldsymbol{U}_{\textrm{true}}}}
\newcommand{\Bstar}{{\boldsymbol{B}^\star}}
\newcommand{\Ttastar}{{\boldsymbol{\Theta}^\star}}

\newcommand{\Ttag}[1]{\boldsymbol{\Theta}_g^{\scalebox{.6}{$(#1)$}}}
\newcommand{\Ttasg}[1]{\boldsymbol{\Theta}_{\setminus g}^{\scalebox{.6}{$(#1)$}}}

\newcommand{\Ttat}[1]{\boldsymbol{\Theta}^{\scalebox{.6}{$(#1)$}}}
\newcommand{\delg}[1]{{\delta^{\scalebox{.6}{$(#1)$}} }}
\newcommand{\rhog}[1]{{\rho^{\scalebox{.6}{$(#1)$}} }}
\newcommand{\psig}[1]{{\psi^{\scalebox{.6}{$(#1)$}} }}

\newcommand{\W}{\boldsymbol{W}}
\newcommand{\U}[2]{\boldsymbol{U}_{#1}^{\scalebox{.6}{$(#2)$}}}
\newcommand{\Ug}{\boldsymbol{U}_{g}}
\newcommand{\Bg}{\boldsymbol{B}_{g}}
\newcommand{\Z}[2]{\boldsymbol{Z}_{#1}^{\scalebox{.6}{$(#2)$}}}
\newcommand{\bZ}{\boldsymbol{Z}}
\newcommand{\bZtilde}{\widetilde{\boldsymbol{Z}}}
\newcommand{\bR}{\boldsymbol{R}}
\newcommand{\B}[2]{\boldsymbol{B}_{#1}^{\scalebox{.6}{$(#2)$}}}
\newcommand{\X}[2]{\boldsymbol{X}_{#1}^{\scalebox{.6}{$(#2)$}}}
\newcommand{\R}[2]{\boldsymbol{R}_{#1}^{\scalebox{.6}{$(#2)$}}}
\newcommand{\G}[2]{\boldsymbol{G}_{#1}^{\scalebox{.6}{$(#2)$}}}
\newcommand{\Xt}{{\boldsymbol{X}_t}}
\newcommand{\yt}{\boldsymbol{y}_t}
\newcommand{\ytt}[1]{\boldsymbol{y}_t^{\scalebox{.6}{$(#1)$}}}
\newcommand{\alpg}[2]{{\alpha}_{#1}^{\scalebox{.6}{$(#2)$}}}

\newcommand{\y}{\boldsymbol{y}}
\newcommand{\graph}{\mathcal{G}}
\newcommand{\Sg}{\mathcal{S}_g}
\renewcommand{\S}{\mathcal{S}}
\newcommand{\Pperp}{\boldsymbol{{P_{{U^{\star}}_\perp}}}}
\newcommand{\sigmax}{{\sigma_{\textrm{max}}^\star}}
\newcommand{\sigmin}{{\sigma_{\textrm{min}}^\star}}

\newcommand{\ceta}{c_{\eta}}
\newcommand{\efin}{\epsilon}
\newcommand{\Df}[2]{\nabla f_{#1}^{\scalebox{.6}{$(#2)$}}}

\newcommand{\minsval}[1]{\sigma_{\textrm{min}}(#1)}
\newcommand{\mineval}[1]{\lambda_{\textrm{min}}(#1)}
\newcommand{\maxsval}[1]{\sigma_{\textrm{max}}(#1)}
\newcommand{\maxeval}[1]{\lambda_{\textrm{max}}(#1)}

\newcommand{\qed}{$\hfill\blacksquare$}
\newcommand{\dspaceX}{\mathcal{X}}
\newcommand{\dspaceY}{\mathcal{Y}}
\newcommand{\Rspace}[1]{\mathbb{R}^{#1}}
\newcommand{\mat}[1]{\boldsymbol{#1}}
\newcommand{\eye}[1]{\boldsymbol{I}_{#1}}
\newcommand{\lr}[1]{\left(#1\right)}
\newcommand{\Ng}{\mathcal{N}_g(\graph)}
\newcommand{\Tta}{\boldsymbol{\Theta}}
\newcommand{\tta}{\boldsymbol{\theta}}

\begin{document}
\title{Beyond Centralization: Provable Communication Efficient Decentralized  Multi-Task Learning}
\author{Donghwa Kang, and Shana Moothedath, \IEEEmembership{Senior Member, IEEE}
\thanks{D. Kang and S. Moothedath are with Electrical and Computer Engineering, Iowa State University. Email: \{dhkang, mshana\}@iastate.edu.}
\thanks{This work is supported by NSF-CAREER 2440455 and NSF 2213069.}}
\maketitle

\begin{abstract}
    Representation learning is a widely adopted framework for learning in data-scarce environments, aiming to extract common features from related tasks.
    While centralized approaches have been extensively studied, decentralized methods remain largely underexplored. 
    {We study decentralized multi-task representation learning in which the features share a low-rank structure.}
    We consider $T$ tasks with $n$ data samples per task, where the observations follow the model $y_{ti}=x_{ti}^\top \theta_t^\star$ for $t=1,...,T$ and $i=1,...,n$. In the decentralized setting, task data is distributed across $L$ nodes and information exchange between the nodes are constrained via a communication network.
    The goal is to recover the $d \times T$ feature matrix $\Tta^\star:=[\theta_1^\star,\theta_2^\star,...,\theta_{T}^\star]$ whose $\textrm{rank}(\Tta^\star)=r\ll\min(d,T)$.
    We propose a new alternating projected gradient descent and minimization algorithm with $\epsilon$-accurate guarantee.
    Comprehensive characterizations of the time, communication, and sample complexities are provided. 
    Importantly, the communication complexity is independent of the target accuracy $\efin$, which significantly reduces communication cost compared to prior methods.
    Numerical simulations validate our theoretical analysis across different dimensions and network topologies, and we identify regimes where the decentralized algorithm can outperform centralized federated learning.
    Under large number of nodes and low-bandwidth, decentralized algorithm can be faster than centralized counterpart. 
\end{abstract}


\section{Introduction}
\label{sec:introduction}
Multi-task representation learning (MTRL) is a machine learning (ML) paradigm for simultaneously learning multiple related models by integrating data from diverse sources. By leveraging shared structure across related yet distinct tasks. MTRL improves the performance of each individual task through collaborative training, even when data for any single task is limited.
Multi-task learning has achieved remarkable success in natural language processing, such as GPT-2 \cite{radford2019language}, GPT-3 \cite{brown2020language}, BERT \cite{devlin2018bert}, and computer vision with models such as CLIP \cite{radford2021learning}. Despite these advances, existing multi-task approaches typically require hundreds or thousands of examples to learn functions that generalize effectively \cite{radford2019language}, posing a significant bottleneck in data-limited applications.
Most of the prior works assume unlimited data samples \cite{du2020few, chen2022active}. In many practical domains, such as medical imaging, drug discovery, fraud detection, and natural (low-resource) language processing, data availability is inherently constrained, limiting applicability of existing methods due to poor sample efficiency. 

Further, most existing works on distributed learning focus on a centralized setting, where data is either stored at a single location or distributed but managed by a central server (fusion center) for aggregation. This is true in the context of multi-task representation learning \cite{du2020few, chen2022active, collins2021exploiting, thekumparampil2021statistically}. 
However, many applications require decentralization, either because data is geographically dispersed or to avoid reliance on a single point of failure. 
Compared to centralized approaches with a fusion center, decentralized optimization has its unique advantages in improving network robustness and improving computation efficiency \cite{sun2020improving, tang2018d, yuan2016convergence, chen2012diffusion, nedic2009distributed}.
Nevertheless, decentralized approaches have received far less attention \cite{moothedath2022fast}, a key reason being that decentralized algorithms have long been treated as a
compromise when the underlying network topology does not allow centralized communication; one has to resort to decentralized communication. So a natural question to ask is: 
{\em Can decentralized algorithms be faster than their centralized counterparts? Are there instances where decentralized MTRL is preferred even when centralized MTRL is feasible?} 

Our goal in this paper is to answer this question and propose an efficient algorithm for decentralized multi-task representation learning.
In this paper, we propose a decentralized MTRL framework, where tasks are distributed across multiple nodes that can only exchange information with their neighbors. Unlike centralized approaches, our framework eliminates the need for a fusion center, making it more robust and scalable in settings with communication constraints. Additionally, we consider a data-scarce regime.
At the core of our approach is a novel decentralized {Diffusion Alternating Gradient Descent and Minimization (Dif-AltGDmin)} algorithm, designed to address the challenges of non-convex, under-sampled, and constrained problems.
We provide convergence guarantees and sample and communication complexities of our approach. 

The main contributions of this paper are threefold.
\begin{itemize}
    \item We develop \emph{Dif-AltGDmin}, a diffusion-based alternating GD and minimization approach to solve the decentralized MTRL problem. In our approach, we learn the shared representation using gradient descent followed by a projection step, and update the task-specific parameters via a closed-form minimization. These local updates are then aggregated across the network through a diffusion mechanism, enabling fully decentralized learning.
    The proposed algorithm is \emph{federated} in nature: only subspace estimates are exchanged rather than raw data.
    \item We establish theoretical guarantees and prove convergence and sample complexity bounds of Dif-AltGDmin algorithm. 
    Our algorithm requires $O(ndrT\cdot\log^2(\max(d,L,\frac{1}{\efin}))$ time in total, and per iteration sample complexity of $O(r^2)$ per task, and $O(dr\cdot\max_g \deg_g\cdot \log Ldr)$ communications at each iteration, where $\max_g \deg_g$ is the maximum node degree in the network.
    The communication complexity is independent of the final accuracy $\efin$ and the number of GD iterations $\Tgd$, a significant improvement compared to the existing approach, Dec-AltGDmin \cite{moothedath2022fast}. (See Table \ref{table: complexities}).
    \item {We validated the effectiveness of the proposed algorithm through numerical simulations by varying network (number of nodes, aggregation steps, and connectivity) and problem (feature dimension, number of tasks, and rank) parameters and compared against baseline methods. Our proposed Dif-AltGDmin algorithm outperforms the decentraized baselines consistently. Further, we demonstrated that the proposed algorithm is effective even with a few aggregation steps, and its execution time is comparable to that of the centralized counterpart. We discuss the regimes where this overtaking can happen.}
\end{itemize}

\begin{table*}[ht]
\caption{Complexity comparison of existing works versus ours. 
The parameters $\kappa$ and $\mu$ are treated as numerical constants, and $\max_g \deg_g$ denotes the maximum node degree.
All methods requires $n \geqslant \max(\log T, \log d, r)$ and $nT\gtrsim (d+T)r(r+\log \frac{1}{\econ})$.
The reported communication complexity for AltGDmin refers to the per-iteration cost with the central server.}
\label{table: complexities}
\centering
\resizebox{0.99\linewidth}{!}{
\begin{tabular}{|l|cc|c|}
\toprule
 & \multicolumn{2}{c|}{\textbf{Time Complexity}} & \textbf{Communication Complexity } \\
 & Initialization & GD & per iteration, per node \\%& $nT \gtrsim$ \\
\midrule
AltGDmin (centralized) \cite{nayer2022fast} 
& $nTdr\cdot \log d$ & $nTdr \log\frac{1}{\efin}$
& $dr \cdot L$ \\
\midrule
Dec-AltGDmin \cite{moothedath2022fast} 
& $nTdr \cdot \log d \cdot \log(\max(L,d,\frac{1}{\efin}))$ 
& $nTdr  \cdot\log\frac{1}{\efin}\cdot \log\max(L,\frac{1}{\efin})$ 
 & $dr \cdot (\max_g \deg_g)$\\
\midrule
\textbf{Dif-AltGDmin (proposed)} 
& $nTdr \cdot \log d \cdot \log(\max (L, d))$ 
& $nTdr \cdot \log\frac{1}{\efin} \cdot \log (\max (L,r))$ 
& $dr \cdot (\max_g \deg_g)$\\
\bottomrule
\end{tabular}
}
\end{table*}
\vspace{-5mm}
\subsection{Related Work}
ti-task representation learning has been extensively explored, with roots traced back to seminal works such as \cite{caruana1997multitask, baxter2000model}. 
Recent works have studied multi-task representation learning under various assumptions. For instance, \cite{du2020few, tripuraneni2021provable, thekumparampil2021statistically, collins2021exploiting, xu2021representation, chen2022active, wang2023improved} focus on learning a shared low-dimensional linear representation across tasks, while other studies \cite{yao2022nlp,zamir2018taskonomy} address empirical approaches to multi-task representation learning. 
Despite the significant success of multi-task representation learning, providing rigorous theoretical guarantees remains a key challenge. Existing theoretical studies either adopt a trace-norm convex relaxation of the original non-convex problem or rely on the assumption that an optimal solution to the non-convex problem is known in the analysis \cite{du2020few,chen2022active, tripuraneni2021provable, knight2024multi}.
The primary focus of these works is to demonstrate the sample efficiency by showcasing the benefit of dimensionality reduction via representation learning.
 Low-rank matrix learning is another related line of literature \cite{nayer2022fast, lrpr_gdmin_2}. In the centralized setting, a convex relaxation via mixed-norm minimization was proposed in \cite{lee2019neurips}, and fast gradient descent–based solutions were studied in \cite{collins2021exploiting, nayer2022fast, lrpr_gdmin_2}.

Our work extends and complements this existing literature in two ways. (i) We present a {\em novel decentralized algorithm} (tasks are allowed to share information only with their neighboring tasks defined via a communication network and there is no fusion center) with provable guarantees and sample efficiency, unlike in existing works \cite{du2020few, tripuraneni2021provable, thekumparampil2021statistically, collins2021exploiting, xu2021representation, chen2022active, wang2023improved} that considered a centralized setting. 
A key challenge in decentralized MTRL is that it is inherently a non-convex problem. While decentralized optimization has been extensively studied in both constrained and unconstrained settings, most existing work focuses on convex formulations, where convergence to a global minimizer can be guaranteed (see survey in \cite{nedic2018network}). In the non-convex setting, such methods can only show convergence to a stationary point. 
Non-convex decentralized optimization has been studied, and algorithms, such as primal-dual methods \cite{hong2017prox, mancino2023decentralized}, gradient tracking methods \cite{xin2021improved,di2016next}, and non-convex extensions of decentralized GD methods \cite{zeng2018nonconvex} are proposed. Some of the works considered non-convex cost functions that are smooth and satisfy the Polyak \L ojasiewicz (PL) condition, which is a generalization of strong convexity to non-convex functions (if a function satisfies the PL condition, then all its stationary points are global minimizers) \cite{xin2021improved}.
A key difference is that, most of the existing works  consider the optimization problem of the form $\min_\theta f(\theta)=\frac{1}{L}\sum_{g=1}^L f_g(\theta)$. In contrast, the decentralized MTRL problem structure is different with a global and a local variable (Eq.~\eqref{eqn: minimization centralized}), requiring new proof techniques to prove estimation guarantees.

Recently, decentralized low-rank matrix recovery problem has been explored in \cite{moothedath2022fully, moothedath2022dec, moothedath2023comparing, moothedath2024decentralized}. Initial works \cite{moothedath2022fully, moothedath2022dec} proposed high-level algorithmic approaches, but these required substantial modifications to provide provable guarantees. Subsequent works \cite{moothedath2023comparing, moothedath2024decentralized} introduced updated algorithms, though without theoretical proofs.
A recent work of ours \cite{moothedath2022fast} introduced the first provable decentralized MTRL algorithm, which relied on an average-consensus procedure for aggregation. However, its $\epsilon$-accuracy guarantees required a number of consensus iterations that scales with $\log(1/\epsilon)$. In contrast, the approach presented in this paper replaces consensus aggregation in \cite{moothedath2022fast} with a diffusion-based update, and the new proof technique eliminates this dependence, resulting in a significant improvement in both efficiency and scalability. 
{The preliminary conference version \cite{DK2025ACC} introduced the diffusion-based algorithm, but presented the algorithm and simulation results without detailed theoretical analysis and proofs.}
(ii)~We consider a data-scarce regime, where the number of samples is smaller than the feature dimension. Our guarantees hold in this setting, unlike prior works \cite{chen2022active, du2020few, tripuraneni2021provable}, which assume that the number of task samples must exceed the feature dimension. Our approach is thus viable for practical applications with large problem sizes and fewer data samples. 

\section{Notations and Problem Formulation}
\noindent\textbf{Notations.} We use bold uppercase letters for matrices, bold lowercase letters for vectors, and regular fonts for scalars. For any positive integer $n$,  $[n]$ denotes the set $\{1,...,n\}$. For any vector $\mat{x}$, $\|\mat{x}\|$ denotes the $\ell_2$-norm of $\mat{x}$. For a matrix, $\|\cdot\|_F$ and $\|\cdot\|$ denote the Frobenius norm and induced $\ell_2$ norm, respectively. The transpose of a vector or matrix is denoted by $^\top$.
For a tall matrix $\mat{M}$, $\mat{M}^\dagger:=(\mat{M}^\top\mat{M})^{-1}\mat{M}^\top$ is a pseudo-inverse of $\mat{M}$, and $\mat{P}_{\mat{M}}:=\mat{M}(\mat{M}^\top\mat{M})^{-1}\mat{M}^\top$ is an orthogonal projection onto the column space of $\mat{M}$.
For an orthonormal matrix $\mat{U}$, the projection operator onto the column space and its orthogonal complement are 
$\mat{P}_{\mat{U}}=\mat{U}\mat{U}^\top$
and  $\mat{P}_{\mat{U}_\perp}:=\eye{}-\mat{U}\mat{U}^\top$.
Given two orthonormal matrices $\mat{U}_1,\ \mat{U}_2\in\Rspace{d\times r}$, the subspace distance is $\SD{\mat{U}_1}{\mat{U}_2}:=\|(\eye{}-\mat{U}_1\mat{U}_1^\top)\mat{U}_2\|$. The notation $a\gtrsim b$ means that $a\geqslant Cb$, for some $C > 1$. 
{We establish our results to hold with probability at least $1-1/d$ or $1-C/d$. Throughout the paper, we refer to such events as occurring \textit{with high probability} (w.h.p.).}
\vspace{1 mm}

\noindent\textbf{Problem Setting.}
Consider $T$ tasks, where each task $t\in[T]$ is associated with a distribution $\mu_t$ over the joint space $\dspaceX\times\dspaceY$. 
Here, $\dspaceX\subseteq \Rspace{d}$ and $\dspaceY\subseteq \Rspace{}$ denote the input and output space. 
For each task $t \in [T]$, we have $n$ number of i.i.d. samples $\{x_{ti},y_{ti}\}_{i=1}^n$ drawn from $\mu_t$.
Each task follows a linear model 
\begin{align*}
    y_{ti}=x_{ti}^\top\tta_t^\star,\quad t=1,...,T,\quad i=1,...,n,
\end{align*}
where $\tta_t^\star\in\Rspace{d}$ is the unknown feature vector.
Stacking the $n$ data samples yields the compact form
\begin{align*}%\label{eqn: representation}
    \yt = \Xt\tta^\star_t, \quad t=1,...,T.
\end{align*}
Multi-task representation learning (MTRL) aims to collectively learn the common underlying representation of the tasks.
In this work, we focus on learning low-dimensional representations, where each feature vector $\tta_t^\star$ is assumed to lie in a common $r$-dimensional subspace of $\Rspace{d}$ with $r\ll\min\{d,T\}$, i.e., the feature matrix $\Tta^\star=[\tta_1^\star, \tta_2^\star, \cdots, \tta_T^\star]\in\Rspace{d\times T}$ has rank $r$.
Let $\Tta^\star\svdeq\Ustar\mat{\Sigma}^\star \mat{V}^{\star\top}$ be the reduced singular value decomposition of $\Ttastar$, where $\Ustar \in \Rspace{d\times r}$, $\mat{V}^\star\in\Rspace{r\times T}$ have orthonormal columns and $\mat{\Sigma}^\star\in\Rspace{r\times r}$ is diagonal with positive singular values on its diagonal. 
Define $\Bstar:=\mat{\Sigma}^\star \mat{V}^{\star\top}=[\mat{b}_1^\star,...,\mat{b}_T^\star]$ so that $\mat{\theta}_t^\star=\Ustar \mat{b}_t^\star$, for $t\in[T]$.
We denote the maximum and minimum singular values of $\mat{\Sigma}^\star$ by $\sigmax$ and $\sigmin$, and the condition number as $\kappa:=\sigmax/\sigmin$.
We consider the high-dimensional, data-scarce setting ($n<d$) where the number of samples per task is significantly smaller than the feature dimension. Within this regime, we seek to reconstruct $\Tta^\star$ using as few samples $n$ as possible. 
%\vspace{2 mm}

\noindent\textbf{Decentralized Setting.}
The $T$ tasks are distributed across $L$ nodes. Node $g$ holds tasks indexed by $\Sg\subset [T]$, with $\S_1,...,\S_L$ partitions $[T]$, i.e., $\cup_{g=1}^L \Sg=[T]$ and $\S_g \cap \S_{j} =\emptyset$ if $g \neq j$, $g,j \in [L]$. 
The network topology is modeled by an undirected connected graph $\graph=(\mathcal{V},\mathcal{E})$, where $\mathcal{V}=[L]$ and $\mathcal{E} \subseteq \mathcal{V} \times \mathcal{V}$ denotes the set of edges.
Each node can only exchange information with its neighbors $\Ng:=\{j:(g,j)\in \mathcal{E}\}$. Let $\textrm{ecc}(g)$ be the eccentricity of node $g\in [L]$, i.e., the maximum shortest-path length from that node to any other node in the network, without traversing the same edge more than once. There is no central coordinating node, thus each node can recover a subset of feature vectors, only for tasks those are contained in $\Sg$, i.e., $\Tta^\star_g:=[\tta_t^\star,\ t\in\Sg]$.
%\vspace{2 mm}

\noindent\textbf{Learning Objective.}
The loss function for the $t^{\rm th}$ task is % given by 
\begin{align*}%\label{eqn:loss}
\ f_t(\mat{U},\mat{b}_t):=\bbE_{(x_{ti},y_{ti})\sim\mu_t}\left[(y_{ti}-x_{ti}^\top\mat{U}\mat{b}_t)^2\right].
\end{align*}
The decentralized MTRL (Dec-MTRL) is formulated as
\begin{align}\label{eqn: minimization centralized}
    &\min_{\substack{\mat{U}\in \bbR^{d \times r}\\\mat{B}\in \bbR^{r \times T}}} f(\mat{U},\mat{B}):=\sum_{g=1}^{L}\sum_{t\in \Sg}\|\yt-\Xt\mat{U}\mat{b}_t\|^2,
\end{align}
where the shared representation $\mat{U}\in\Rspace{d\times r}$ is an orthonormal matrix whose columns span $\Ustar$ and the task-specific coefficients $\mat{B}=[\mat{b}_1, \ldots, \mat{b}_T]\in\Rspace{r\times T}$.
This factorization exploits the low-rank structure of $\Tta^\star$, reducing computational complexity by learning only $dr+Tr$ parameters instead of $dT$, which is a substantial saving since $r \ll \min\{d, T\}$. 
\vspace{2 mm}

\noindent\textbf{Assumptions.}
Since no $\yt$ is a function of the whole matrix $\Tta^\star$, recovering $\Tta^\star$ from local task data requires an additional condition to ensure each local dataset contains a sufficiently rich representation of the common subspace.
The following assumption guarantees this property.
\begin{assumption}\label{assumption: right-incoherence}
    For $t\in [T]$, $\|\mat{b}_t\|^2\leqslant \mu^2 \frac{r}{T} \sigmax^2$, where $\mu>1$. The same bound holds for each $\mat{x}_t^\star$ since $\|\mat{x}_t^\star\|^2=\|\mat{b}_t^\star\|^2$.
\end{assumption}
Assumption~\ref{assumption: right-incoherence} referred to as {\em incoherence} originated in \cite{candes2008exact} and has been widely adopted in low-rank matrix learning \cite{nayer2022fast, lee2019neurips} and multi-task representation learning  \cite{lin2024fast,tripuraneni2021provable}.
Additionally, we employ the following two commonly used assumptions.
\begin{assumption}\label{assumption: iid}
    Each entry of $\Xt$ is independently and identically distributed (i.i.d.) standard Gaussian random variable.
\end{assumption}
Assumption \ref{assumption: iid} enables the use of probabilistic concentration bounds and is crucial for reliable subspace recovery in a data-scarce regime. 
It is commonly assumed in most works on MTRL with theoretical guarantees \cite{tripuraneni2021provable, collins2021exploiting, OurICML}, and potential
extensions beyond this assumption are part of our future work.
\begin{assumption}\label{assumption: connected graph}
    The graph, denoted by $\graph$ is undirected and connected, meaning any two nodes are connected by a path.
\end{assumption}
Assumption \ref{assumption: connected graph} ensures that local information can propagate, allowing all nodes to reach consensus. This is a standard assumption in decentralized control, estimation, and learning \cite{olshevsky2009convergence, olfati2004Consensus}. We do not assume a fully connected network.

We highlight three main challenges in solving Dec-MTRL and outline our approach to address them below. 
\begin{enumerate}
    \item A key challenge arises from the coupling of parameters $\mat{U}$ and $\mat{B}$. 
    Further, the cost function has many global minima, including all pairs
    of matrices $(\Ustar\mat{Q}^\top, \mat{Q}\Bstar)$, where $\mat{Q}\in \bbR^{r \times r}$ is orthonormal, eliminating the possibility of learning the ground truth parameters $(\Ustar, \Bstar)$. 
    Thus we aim to learn the shared {\em representation}, i.e., column space of $\Ustar$, and the task-specific parameters, $\mat{b}_t$ for $t\in [T]$, to minimize $f(\mat{U},\mat{B})$. We adopt alternating projected gradient descent and minimization (AltGDmin) \cite{nayer2022fast}, where $\mat{B}$ is updated via least-squares minimization with $\mat{U}$ fixed, and $\mat{U}$ is updated via projected (orthonormal) gradient descent with $\mat{B}$ fixed. See Section~\ref{ssc: Dec-AltGDmin}.
    
    \item In the decentralized setting, the tasks are distributed among the nodes, and each node $g \in [L]$ observes only a subset $\Sg\subseteq [T]$ of the tasks. 
    Communication between nodes is constrained within their neighboring nodes, defined by the network.
    Hence, node $g$ can only evaluate its local cost function
    $f_g(\Ug,\Bg)=\sum_{t\in\Sg}\|\yt-\Xt\Ug\mat{b}_t\|^2$, from which it can compute local gradient $\nabla_{\boldsymbol{U}}f_g(\mat{U}_g,\mat{B}_g)$.
    The update of $\Ug$ requires collaboration with neighboring nodes, and $\Bg$ is subsequently updated conditioned on $\Ug$.
    Standard decentralized optimization methods such as \cite{yuan2016convergence, nedic2009distributed} are not directly applicable to Dec-MTRL, due to the non-convexity. 
    Building on \cite{moothedath2022fast}, we propose diffusion-based aggregation, 
    and our refined analysis provides a tighter bound on the communication complexity. See Sections~\ref{sssc: AvgCons} and \ref{ssc: Dec-AltGDmin}.
    
    \item Since the cost function in Eq.~\eqref{eqn: minimization centralized} is non-convex, a good initial estimate of $\mat{U}$ close to $\Ustar$ is essential for convergence within desired accuracy. 
    This is further complicated by the need for decentralized initialization, which necessitates inter-node consistency among the nodes. 
    To address this challenge, we adapt the decentralized noisy power method from \cite{hardt2014noisy} and employ the decentralized truncated spectral initialization, proposed in our previous work \cite{moothedath2022fast}, with a slight modification.
    Building on \cite{moothedath2022fast}, we use broadcasting to ensure exact consensus, i.e., $\|\Ug-\mat{U}_{g'}\|_F=0$. Our new analysis provides a tighter bound on communication complexity (Section~\ref{sssc: initialization}).
\end{enumerate}
Additionally, we employ a sample-splitting technique to ensure independence between the gradient updates and minimization step in each iteration of the algorithm.
In the next section, we provide the details of the proposed algorithm. 

\section{A New Algorithm for Decentralized MTRL}\label{sec: algorithm}
We begin with the preliminary background and then introduce the proposed algorithm for solving Dec-MTRL.
%\vspace{-5mm}
\subsection{Preliminaries}
\subsubsection{Agreement Protocol}\label{sssc: AvgCons}
\begin{algorithm}[t]
\caption{Agreement Algorithm (\textsc{Agree})}
\label{alg:AvgCons}
\begin{algorithmic}[1]
\State \textbf{Input:} $\Z{g}{\mathrm{in}},\ \forall g\in[L],\ \Tcon,\ \graph$
\State Initialize $\Z{g}{0} \leftarrow \Z{g}{\mathrm{in}}$
\For{$\tau = 0$ \textbf{to} $\Tcon-1$, for each $g\in[L]$}
        \State $\Z{g}{\tau+1} \leftarrow \Z{g}{\tau} + \sum_{j\in \Ng} \W_{gj}\left(\Z{j}{\tau}-\Z{g}{\tau}\right)$
        
\EndFor
\State \textbf{Output:} $\Z{g}{\mathrm{out}} \leftarrow \Z{g}{\Tcon}$
\end{algorithmic}
\end{algorithm}
To coordinate the estimation of the shared representation $\mat{U}$, nodes exchange information with their neighbors
and perform weighted averaging
(Line~\ref{step:cons} in Algorithm~\ref{alg:dec-altgd}).
This operation functions as a general agreement mechanism: all nodes asymptotically converge to the global average of the initial values.
Such updates encompass the average \textit{consensus} employed in our earlier works \cite{moothedath2022fast, moothedath2022dec} and diffusion-type updates depending on the exchanged variables. 
We formalize this in Algorithm~\ref{alg:dec-altgd}.
Let $\W\in\Rspace{L\times L}$ be a doubly stochastic weight matrix, and
define the connectivity $\gamma(\W):=\max(|\lambda_2(\W)|,\ |\lambda_L(\W)|)$.
Under standard connectivity assumptions, 
it holds that $\lim_{k\rightarrow\infty}\W^k\rightarrow\tfrac{1}{L}\mat{1}\mat{1}^\top$, where $\gamma(\W)<1$ governs the speed of the convergence.
Repeating finite times of \textsc{Agree} yields an approximation of the global average. 
The following result quantifies the desired error bound and the required number of agreement rounds. 
\begin{proposition}\label{prop: avgcons}(\cite{olshevsky2009convergence})
    Consider the agreement algorithm in Algorithm \ref{alg:AvgCons} with doubly stochastic weight matrix $\W$. Let $z_\mathrm{true}:=\frac{1}{L}\sum_{g=1}^L z^{\scalebox{.6}{(in)}}_{g}$ be the true average of the initial values $z^{\scalebox{.6}{(in)}}_{g}$ across $L$ nodes. For any $\econ<1$, if the graph is connected and if $\Tcon\geqslant \frac{1}{\log(1/\gamma(\W))}\log(L/\econ)$, then the outputs after $\Tcon$ iterations satisfy
   
      $\qquad\max_g |z^{\scalebox{.6}{(out)}}_{g}-z_\mathrm{true}|\leqslant \econ\max_g|z^{\scalebox{.6}{(in)}}_{g}-z_\mathrm{true}|.$\hfill$\Box$
\end{proposition}

Proposition \ref{prop: avgcons} is stated for scalar consensus, but it naturally extends to matrix valued variables (See Proposition \ref{prop: ConsErr_prop}).
Rewriting $\Tcon$ expression in terms of $\gamma(\W)$ and fixing $L$ and $\Tcon$ gives the connectivity requirement for desired accuracy 
\begin{align}\label{eqn: connectivity}
    \gamma(\W)\leqslant\exp\left(-C\frac{\log(L/\econ)}{\Tcon}\right).
\end{align}
This inequality highlights the trade-off between accuracy and communication: achieving higher accuracy necessitates stronger network connectivity, i.e., smaller $\gamma(\W)$.

%\vspace{2 mm}
\begin{algorithm}[t]
\caption{Modified Decentralized Spectral Initialization}
\label{alg:dec-init}
\begin{algorithmic}[1]
\State \textbf{Input:} $\{\Xt,\yt\}_{t\in\Sg},\ \graph,\ \kappa,\mu,n,T,\Tpm,\Tconinit$
\State Let $\yt \equiv \ytt{00},\ \Xt \equiv \X{t}{00}$
\State $\alpg{g}{\rm in}\leftarrow 9\kappa^2\mu^2\frac{L}{nT}\sum_{t\in\Sg}\sum_{i=1}^n y_{ti}^2$
\State $\alpha_g \leftarrow \AvgCons(\alpg{g'}{\rm in},\ g'\in[L],\ \graph,\ \Tconinit)$
\State Let $\yt \equiv \ytt{0},\ \Xt \equiv \X{t}{0}$
\State $\y_{t,trnc}:=\yt\circ\indic_{\{y_{ti}^2\leqslant\alpha_g\}}$
\State $\Ttag{0}=\Big[\tfrac{1}{n}\Xt^\top \y_{t,trnc},\ t\in\Sg\Big]$
\State Generate $\Utilde{g}{\rm in}$ with i.i.d.\ standard Gaussian entries (same seed for all $g$)
\State $\Utilde{g}{\rm in}\qreq \U{g}{0}\R{g}{0}$, so $\U{g}{0}=\Utilde{g}{\rm in}(\R{g}{0})^{-1}$
\For{$\tau=1$ \textbf{to} $\Tpm$, each $g\in[L]$}  
    \State $\Utilde{g}{\rm in}\leftarrow \Ttag{0}{\Ttag{0}}^\top \U{g}{\tau-1}$
    \State $\Utilde{g}{\tau}\leftarrow \AvgCons(\Utilde{g'}{\rm in},\ \graph,\ \Tconinit)$
    \State $\Utilde{g}{\tau}\qreq \U{g}{\tau}\R{g}{\tau}$, and get $\U{g}{\tau}$
    \State $\U{g=1}{\tau}=\U{g=1}{\tau},\quad \U{g}{\tau}=\mathbf{0}$ for $g\neq 1$\label{alg2: zeroing}
    \For{ {$\tau_{\textrm{con}}=1$ \textbf{to} $\textrm{ecc}(1)$, each $g\in[L]$}} \label{alg2: broadcast - begin}
    \State { \textbf{If $\U{g}{\tau}\neq 0$}: Send $\U{g}{\tau}$ to $j\in \mathcal{N}_g(\mathcal{G})$}
    \State { \textbf{If received $\U{j}{\tau}\neq 0$}: $\U{g}{\tau}\leftarrow \U{j}{\tau}$ }\label{alg2:broadcast}
    \EndFor\label{alg2: broadcast - end}
\EndFor
\State \textbf{Output:} $\U{g}{\Tpm}$
\end{algorithmic}
\end{algorithm}
\subsubsection{Decentralized Spectral Initialization}\label{sssc: initialization}
Since the cost function in Eq.~\eqref{eqn: minimization centralized} is non-convex, a proper initialization is needed. 
We adopt the {\em decentralized truncated spectral initialization} first proposed in \cite{moothedath2022fast} with slight modification. The spectral initialization aims at computing an initial estimate $\U{g}{0}$ close enough to $\Ustar$ while maintaining inter-node consistency, i.e., keeping $\|\U{g}{0}-\U{g'}{0}\|_F$ small.
In particular, for inter-node consistency, the algorithm uses agreement protocol after line~\ref{alg2: zeroing} to propagate $\U{1}{0}$ to all nodes.
Our approach replaces this step (lines~\ref{alg2: broadcast - begin}–\ref{alg2: broadcast - end}) with a direct broadcast of node 1’s exact estimate.
This modification guarantees $\delg{0}$-accurate initial estimate of the shared representation matrix $\Ustar$ and exact node-wise consistency, as formalized in Proposition \ref{prop: initialization}.
In addition, it quantifies the sample and iteration complexities.
The decentralized spectral initialization is summarized in Algorithm \ref{alg:dec-init}.
For full details, see Theorem 4.4 of \cite{moothedath2022fast} and its proof. 

\begin{proposition}\label{prop: initialization}
    (Modified Decentralized Spectral Initialization)     
    Consider the decentralized truncated spectral initialization algorithm in Algorithm~\ref{alg:dec-init} and suppose that Assumptions \ref{assumption: right-incoherence}-\ref{assumption: connected graph} hold. Pick $\delg{0}$.  
    If $\Tconinit\geqslant \max C\frac{1}{\log1/\gamma(\W)}(\log L+\log d+\log\kappa+\log(1/\delg{0}))$, $\Tpm\geqslant C\kappa^2\log(d/\delg{0})$, and $nT\gtrsim\kappa^4\mu^2(d+T)\frac{r}{\delg{0}^2},$ 
    then w.p. at least $1-1/d$,
    \begin{enumerate}
        \item $\max_{g\neq g'}\|\U{g}{\Tpm}-\U{g'}{\Tpm}\|_F=0$
        \item $\max_{g\neq g'}\|\Pperp(\U{g}{\Tpm}-\U{g'}{\Tpm})\|_F=0$.
        \item $\SD{\U{g}{\Tpm}}{\Ustar}\leqslant \delg{0}$.
    \end{enumerate}
    
   \begin{proof}
      Broadcasting in lines \ref{alg2: broadcast - begin}-\ref{alg2: broadcast - end} ensures $\U{g}{\tau}=\U{1}{\tau}$ for all $g\in [L]$ and $\tau\in [\Tpm]$. Thus, $\|\U{g}{\tau}-\U{1}{\tau}\|_F=0$ for all $\tau=1, \ldots, \Tpm$ and parts 1) and 2) hold.
    
        Part 3) follows similar steps in Section VI of \cite{moothedath2022fast} with slight modifications. The main idea is to bound the subspace distance in every PM round $\tau$. For all $\tau\in [\Tpm]$, we have
        \begin{align*}
            \SD{\U{g}{\tau}}{\Ustar}&\leqslant \SD{\U{g}{\tau}}{\U{1}{\tau}}+\SD{\U{1}{\tau}}{\Ustar}\\
            &\leqslant \|\U{g}{\tau}-\U{1}{\tau}\|_F+\SD{\U{1}{\tau}}{\Ustar}.
        \end{align*}
        From 1), the first term is zero. 
        Using triangular inequality, the second term breaks into
        \begin{align*}
            \SD{\U{1}{\tau}}{\Ustar}\leqslant\SD{\Utrue}{\Ustar}+\SD{\U{1}{\tau}}{\Utrue}.
        \end{align*}
        For the bound of $\SD{\Utrue}{\Ustar}$, refer to Lemma 6.2 of \cite{moothedath2022fast}, which guarantees $\SD{\Utrue}{\Ustar}\leqslant 0.5\delg{0}$ w.h.p if $nT\gtrsim \kappa^4\mu^2(d+T)r/\tilde{\delta}_0^2$ for some  $\tilde{\delta}_0^2<0.1$.
        To bound $\SD{\U{1}{\tau}}{\Utrue}$, we use noisy power method (Proposition 6.3 \cite{moothedath2022fast}), 
        which requires 
        \begin{align*}
            \|\G{1}{\tau}\|< \min\left(0.5\delg{0},\tfrac{1}{d^{1/c_1}\sqrt{r}}\right)0.8\sigmin^2.
        \end{align*}
        where $\G{1}{\tau}:=\Utilde{1}{\tau}-(\Tta_0)(\Tta_0)^\top\U{1}{\tau-1}$.
        We show that this bound relies on $\econ$ and obtain the achievable bound for $\|\G{1}{\tau}\|_F$ to derive required $\Tconinit$ to satisfy this required bound.
        By adding and subtracting $\sum_g(\Tta_0)_g{(\Tta_0)_g}^\top\U{g}{\tau-1}$, we decompose $\G{1}{\tau} = \ConsErr{1}{\tau}+\UconsErr{1}{\tau}$, where
        \begin{align*}
            \ConsErr{1}{\tau}&:=\Utilde{1}{\tau}-\sum_g(\Tta_0)_g{(\Tta_0)_g}^\top\U{g}{\tau-1} \\
            \UconsErr{1}{\tau}&:=\sum_g(\Tta_0)_g{(\Tta_0)_g}^\top(\U{g}{\tau-1}-\U{1}{\tau-1}).
        \end{align*}
        Part 1) yields $\|\UconsErr{1}{\tau}\|_F=0$. 
        To bound $\ConsErr{1}{\tau}$, we follow Section VI. B of \cite{moothedath2022fast}, which gives $\|\ConsErr{1}{\tau}\|_F\leqslant 1.1L\sqrt{r}\econ \sigmax^2$ if $\Tconinit\geqslant C\frac{1}{\log(1/\gamma(\W))}\log(\frac{L}{\econ})$ and $nT\gtrsim \kappa^2\mu^2(d+T)r$. 
        Hence, the bound for noisy PM holds if
        \begin{align*}
            \econ=0.8\min\left(\tfrac{\delg{0}}{2\kappa^2L\sqrt{r}},\tfrac{1}{\kappa^2 dL}\right),
        \end{align*}
        meaning that we need to set
        \begin{align*}
            \Tconinit=C\frac{\log L+\log d+\log \kappa +\log (1/\delg{0})}{\log(1/\gamma(\W))}.
        \end{align*}
        For $\Tpm$ expression, see Section VI. C of \cite{moothedath2022fast}, which gives 
            $\Tpm\geqslant C\kappa^2\log (\frac{d}{\delg{0}})$.
        This completes the proof. 
   \end{proof}
\end{proposition}

% \vspace{-3mm}
\subsection{Proposed Algorithm: Diffusion-based Alternating Projected GD and Minimization (Dif-AltGDmin)}\label{ssc: Dec-AltGDmin}
\begin{algorithm}[t]
\caption{Diffusion-based Alternating Gradient Descent and Minimization (Dif-AltGDmin)}
\label{alg:dec-altgd}
\begin{algorithmic}[1]
\State \textbf{Input:} $\Xt,\ \yt,\ t\in\Sg,\ g\in[L],\ \graph$
\State \textbf{Output:} $\U{g}{\Tgd},\ \B{g}{\Tgd},\ \Ttag{\Tgd}=\U{g}{\Tgd}\B{g}{\Tgd}$
\State \textbf{Parameters:} $\eta,\ \Tcongd,\ \Tgd$
\State \textbf{Sample-split:} Partition $\Xt,\ \yt$ into $2\Tgd+2$ disjoint sets $\X{\tau}{\ell},\ \ytt{\ell},\ \ell=00,0,1,\dots,2\Tgd$
\State \textbf{Initialization:} Run Algorithm~\ref{alg:dec-init} to get $\U{g}{0}\leftarrow \U{g}{\Tpm}$
\For{$\tau=1$ \textbf{to} $\Tgd$, for each $g\in[L]$}
    \State Let $\yt\equiv\ytt{\tau},\ \Xt\equiv\X{\tau}{\tau}$
    \State $\bt{\tau}\leftarrow (\Xt\U{g}{\tau-1})^\dagger \yt \quad \forall t\in\Sg$
    \State $\ttat{\tau}\leftarrow \U{g}{\tau-1}\bt{\tau} \quad \forall t\in\Sg$
    \State Let $\yt\equiv\ytt{\tau+\Tgd},\ \Xt\equiv\X{\tau}{\tau+\Tgd}$
    \State $\Df{g}{\tau}\leftarrow \sum_{t\in\Sg}\Xt^\top(\Xt\U{g}{\tau-1}\bt{\tau}-\yt)\bt{\tau}^\top$
    \State {\bf Local update:} $\Ubreve{g}{\tau}\leftarrow \U{g}{\tau-1}-\eta L\Df{g}{\tau}$ \label{line:cons-U}
    \State {\bf Diffusion:} $\Utilde{g}{\tau}\leftarrow \AvgCons(\Ubreve{g}{\tau},\ \graph,\ \Tcongd)$ \label{step:cons}
    \State \textbf{Projection:} $\Utilde{g}{\tau}\qreq \U{g}{\tau}\R{g}{\tau}$ and $\U{g}{\tau}\leftarrow \Utilde{g}{\tau}{\R{g}{\tau}}^{-1}$\label{line:proj}
\EndFor
\State \textbf{Output:} $\U{g}{\Tgd},\ \B{g}{\Tgd}=\begin{bmatrix}
    \bt{\Tgd},\ t\in[T]
\end{bmatrix},\ \Ttag{\Tgd}=\U{g}{\Tgd}\B{g}{\Tgd}$
\end{algorithmic}
\end{algorithm}
We present the proposed Dif-AltGDmin algorithm in Algorithm~\ref{alg:dec-altgd}.
Starting from the initialization, each round consists of two main steps: (i) minimization step for $\Bg$, (ii) {diffusion-based projected gradient descent for $\Ug$.}

{\bf Step 1. $\mat{B}$ update:} 
Consider the $g$-th node at iteration $\tau$.
Given the most recent estimate 
$\U{g}{\tau-1}$, the node computes $\B{g}{\tau}=\argmin_{\Bg} f_g(\U{g}{\tau-1},\Bg)$.
This minimization decouples to a set of column-wise least square problems:
\begin{align*}%\label{eq:b_update}
\bt{\tau}=(\Xt\U{g}{\tau-1})^\dagger \yt,\quad\forall t\in\Sg,
\end{align*}
and collecting the columns yields $\B{g}{\tau}=[\bt{\tau},\ t\in\Sg]$.

 {\bf Step 2. $\mat{U}$ update:}
Once node $g$ obtains $\B{g}{\tau}$, it performs a projected gradient descent step to update $\U{g}{\tau}$. 
Specifically, the local gradient is computed as 
$\Df{g}{\tau}:=\nabla_{\mat{U}}f_g(\U{g}{\tau-1},\B{g}{\tau})$
and a \textit{local update} step is carried out as
\begin{align}\label{eq:U_update}
\Ubreve{g}{\tau}\leftarrow \U{g}{\tau-1}-\eta L\Df{g}{\tau}.
\end{align}
Recall that the global objective in Eq.~\eqref{eqn: minimization centralized} is the sum of local costs.
Accordingly, the centralized descent takes the form 
\begin{align*}
    \Uhat{\tau}&:=\frac{1}{L}\sum_{g=1}^L\lr{\U{g}{\tau-1}-\eta L\Df{g}{\tau}}=\Ubar{\tau-1}-\eta\gradU{\tau},
\end{align*}
where $\Ubar{\tau}:=\frac{1}{L}\sum_{g=1}^L\U{g}{\tau}$ denotes the true average and $\gradU{\tau}:=\sum_{g=1}^L\Df{g}{\tau}$ is the aggregated gradient.
Note that $\Uhat{\tau}$ is not available to local nodes.
In the decentralized setting, each node exchanges the local estimate $\Ubreve{g}{\tau}$ with neighboring nodes and applies the agreement protocol, producing an \textit{averaged update} $\Utilde{g}{\tau}$.
This asymptotically approximates the centralized update $\Uhat{\tau}$.
Finally, each node performs a local QR decomposition $\Utilde{g}{\tau}\qreq\U{g}{\tau}\R{g}{\tau}$ and sets the \textit{projected update} as $\U{g}{\tau}=\Utilde{g}{\tau}(\R{g}{\tau})^{-1}$.
Note that only $\Utilde{g}{\tau}$ is exchanged among nodes, while task-specific parameter $\bt{\tau}$ remain local (federation).
{Overall, the proposed decentralized update rule approximates the centralized gradient descent step 
${ \U{}{\tau+1}=\U{}{\tau}-\eta \nabla_U(f(\U{}{\tau},\B{}{\tau}))}$ 
which would be obtained if all data were available at a fusion server.}

%

% \vspace{-8 mm}
\section{Main Result and Discussion}\label{sec:main result}
In this section, we present our main theoretical guarantee, which establishes the accuracy of the proposed Dif-AltGDmin algorithm.
We begin by presenting our main result, and then analyze the sample, time, and communication complexities.
\vspace{-2 mm}
\subsection{Main Result}\label{ssc: main result}
Theorem~\ref{thm: main} states that, under standard incoherence, i.i.d., and network connectivity assumptions, our method recovers the subspace of the common representation matrix $\Ustar$ and the ground-truth task parameters $\tta_t^\star$ within a prescribed error tolerance. 
It also presents the required sample size, iteration complexity, and communication requirement to achieve the desired accuracy. Proof of Theorem \ref{thm: main} is presented in Section~\ref{sec: proof of main thm}.
\begin{theorem}\label{thm: main}
Suppose that Assumptions \ref{assumption: right-incoherence}-\ref{assumption: connected graph} hold. 
Consider the outputs $\U{g}{\Tgd}$ and $\Ttag{\Tgd}$ for all $g\in[L]$ of Algorithm \ref{alg:dec-altgd} initialized with Algorithm \ref{alg:dec-init}. Pick $\efin<1$ and let $\eta=0.4/n\sigmax^2$.
Assume that 
\begin{enumerate}[label=\alph*)]
    \item $\Tpm=C\kappa^2(\log d+\log \kappa)$ and\\
        $\Tconinit= C\frac{1}{\log(1/\gamma(\W))}(\log L+\log d+\log r+\log\kappa)$;
    \item $\Tgd=C\kappa^2\log(1/\efin)$ and\\
        $\Tcongd= C\frac{1}{\log(1/\gamma(\W))}(\log L+\log r+\log\kappa)$;
    \item $nT \geqslant C\kappa^6\mu^2(d+T)r(\kappa^2r+\log(1/\efin))$.
\end{enumerate}
Then, with probability at least $1-1/d$,
\begin{enumerate}
    \item The task parameters for all $t\in \mathcal{S}_g$ and $g\in[L]$, are recovered up to an $\efin$-error, i.e., $$\|\ttat{\Tgd}-\tta_t^\star\|\leqslant1.4\efin\|\tta_t^\star\|;$$
    \item The subspace distance, for all $g\in[L]$, satisfies
    % \vspace{5mm}
    
    $\qquad \qquad \qquad \SD{\U{g}{\Tgd}}{\Ustar}\leqslant\efin.$\hfill$\Box$
\end{enumerate}

\end{theorem}
\subsection{Time, Communication and Sample Complexities}
\label{ssc: complexities}
\noindent\textbf{Time complexity.} The time complexity of Dif-AltGDmin is 
\begin{align}\label{eqn: time-complexity}
    \mathcal{\tau}_{\mathrm{time}}
    =(\Tconinit\cdot \Tpm)\varpi_{\mathrm{init}}+(\Tcongd\cdot\Tgd)\varpi_{\mathrm{gd}}
\end{align}
where $\varpi_{\mathrm{init}}$ and $\varpi_{\mathrm{gd}}$ denote the per-step computational costs for initialization and Dif-AltGDmin, respectively.
For Dif-AltGDmin, each iteration involves
(i) task-wise least-square updates with cost $O(ndr\cdot |\Sg|+nr^2\cdot|\Sg|)$; $\bt{\tau}=(\Xt\U{g}{\tau-1})^\dagger \yt$ requires $ndr\cdot |\Sg|$ for computing $\Xt\U{g}{\tau-1}$ and $nr^2\cdot|\Sg|$ for least-square,
(ii) the gradient evaluation with cost $O(ndr\cdot|\Sg|)$; $\Df{g}{\tau}=\sum_{t\in\Sg}\Xt^\top( \Xt\U{g}{\tau-1}\bt{\tau}-\yt)\bt{\tau}^\top$, and QR factorization with cost $O(dr^2)$ per node.
Since $\sum_{g=1}^L|\Sg|=T$, the aggregated computational cost over all $L$ nodes for Dif-AltGDmin step is $\varpi_{\mathrm{gd}}=O(ndrT)$.
Similarly, each PM round in the initialization phase requires $\varpi_{init}=ndrT$.
Substituting the above and parts a) and b) of Theorem \ref{thm: main} into Eq.~\eqref{eqn: time-complexity}, the total time complexity is roughly
\begin{align*}
    \mathcal{\tau}_{\mathrm{time}}&=\underbrace{(\Tconinit\cdot \Tpm)\varpi_{\mathrm{init}}}_{\mathcal{\tau}_\mathrm{init}}+\underbrace{(\Tcongd\cdot\Tgd)\varpi_{\mathrm{gd}}}_{\mathcal{\tau}_\mathrm{gd}}\\
    &\approx C\kappa^2\frac{\max(\log^2d,\log^2\kappa,\log^2L,\log^2(1/\efin))}{\log(1/\gamma(\W))}\cdot ndrT.
\end{align*}

To make the improvement more transparent, we separate the time complexities for initialization and GD steps
\begin{align*}
    \mathcal{\tau}_\mathrm{init}
    &\approx \frac{C\kappa^2\log d\cdot \max(\log d,\log\kappa,\log L)}{\log(1/\gamma(\W))} ndrT\\
    \mathcal{\tau}_\mathrm{gd}
    &\approx \frac{C\kappa^2\log(\frac{1}{\efin})\cdot \max(\log L,\log r,\log\kappa)}{\log(1/\gamma(\W))} ndrT.
\end{align*}
In contrast, the corresponding time complexities in \cite{moothedath2022fast} are
\begin{align*}
    \mathcal{\tau}_\mathrm{init}^*
    &\approx \frac{C\kappa^4\log d\cdot \max(\log d,\log \kappa,\log L,\log (\frac{1}{\efin}))}{\log(1/\gamma(\W))} ndrT\\
    \mathcal{\tau}_\mathrm{gd}^*
    &\approx \frac{C\kappa^4\log(\frac{1}{\efin})\cdot \max(\log(\frac{1}{\efin}),\log L,\log\kappa)}{\log(1/\gamma(\W))} ndrT.
\end{align*}
Hence, our algorithm achieves:
\begin{enumerate}
    \item Improved scaling in $\kappa$: both complexities involve $\kappa^2$ rather than $\kappa^4$.
    \item Less dependence on $\log(1/\efin)$: we avoid $\log(1/\efin)$ in $\mathcal{\tau}_\mathrm{init}$ and reduce $\log^2(1/\efin)$ to $\log(1/\efin)$ in $\mathcal{\tau}_\mathrm{gd}$.
    This is because our $\Tcongd$ does not depend on $\log(1/\efin)$, which is a significant improvement over the approach in \cite{moothedath2022fast}.
    \item Less dependence on $d$: $\mathcal{\tau}_\mathrm{gd}$ does not involve $\log d$ compared to \cite{moothedath2022fast}.
\end{enumerate}
 We remark that, since all nodes operate in parallel, the actual execution time scales with the node with the heaviest load, i.e., $\mathcal{\tau}_\mathrm{gd}
    \approx C\kappa^2\frac{\log(1/\efin)\cdot \max(\log L,\log r,\log\kappa)}{\log(1/\gamma(\W))}\cdot ndr\max_g|\Sg|$.

\noindent\textbf{Communication complexity.}
At each communication round, each node exchanges a $d\times r$ matrix with its neighbors. 
Since node $g$ has degree $\deg_g$, the per-round communication cost is $O(dr\cdot \max_g\deg_g\cdot L)$.
Over all rounds of initialization and Dif-AltGDmin, the total communication complexity is roughly
\begin{align*}
    &\mathcal{\tau}_{\mathrm{comm}}=(\Tconinit \Tpm+ \Tcongd \Tgd)\cdot \lr{drL(\max_g \deg_g)}\\
    &\approx drL(\max_g \deg_g)\frac{C\kappa^2\max(\log^2d,\log^2\kappa,\log^2L,\log^2(\frac{1}{\efin}))}{\log(1/\gamma(\W))}.
\end{align*}

\noindent\textbf{Sample complexity.} Consider the lower bound of $nT$ specified in Theorem \ref{thm: main}.
Suppose $d\approx T$ for simplicity, then our method requires roughly only $O(r^2)$ samples per task (ignoring logarithmic factors and constants).
In contrast, a naive approach without the low-rank assumption would need at least $n\geqslant d$ samples per task, since each $\tta_t$ should be recovered via $\tta_t=\Xt^\dagger\yt$.
Therefore, our framework achieves reliable recovery with significantly fewer samples, making it particularly effective in data-scarce regime, where $n<d$.
\begin{remark}\label{remark: Tcon}
    The sample complexity per task remains of the same order $O(r^2)$ as in our earlier work \cite{moothedath2022fast} for achieving a target accuracy $\efin$. 
    In contrast, our communication requirement is significantly lower: \cite{moothedath2022fast} requires 
    $\Tconinit=C\frac{1}{\log1/\gamma(\W)}(\log L+\log \Tgd+\log(1/\efin)+\log d+\log\kappa)$ 
    and 
    $\Tcongd=C\frac{1}{\log1/\gamma(\W)}(\log L+\log \Tgd+\log(1/\efin)+\log\kappa)$.
    We rigorously prove this communication saving effect in Section~\ref{sec: proof of main thm} and corroborate the results through simulations presented in Section ~\ref{sec: simulations}.
\end{remark}

\section{Preliminaries}\label{sec: prelim}
In this section, we define the notations and present preliminary results necessary for our analysis.

\noindent{\bf Definitions:}
Consider local estimates of shared representation $\U{g}{\tau}$ and local gradients $\Df{g}{\tau}$, for $g\in[L]$.
Define the true average $\Ubar{\tau}$ and the aggregated gradient $\gradU{\tau}$ as  
\begin{align*}
    \Ubar{\tau}:=\frac{1}{L}\sum_{g=1}^L \U{g}{\tau}\quad\mathrm{and}\quad 
    \gradU{\tau}:=\sum_{g=1}^L\Df{g}{\tau}.
\end{align*}
Based on this, we define the true average $\Uhat{\tau}$ that the \textsc{Agree} algorithm aims to approximate and the \textit{consensus error} between the true average and the actual output $\Utilde{g}{\tau}$
\begin{align*}
    &\Uhat{\tau}:=\frac{1}{L}\sum_{g=1}^L(\U{g}{\tau-1}-\eta L\Df{g}{\tau})=\Ubar{\tau-1}-\eta \gradU{\tau},\\
    &\ConsErr{g}{\tau}:=\Utilde{g}{\tau}-\Uhat{\tau}.
\end{align*}
The inter-node consensus error, i.e., $\mat{U}$ estimation error between nodes $g$ and $g'$, is defined as
\begin{align*}
    \UconsErr{g,g'}{\tau}:=\U{g}{\tau}-\U{g'}{\tau}.
\end{align*}
and define $\Pperp:=I-\Ustar\Ustar^\top$.
We also define the deviation of the aggregated gradient from its expected value
\begin{align*}
    \Err{\tau}:=\bbE[\gradU{\tau}]-\gradU{\tau}.
\end{align*}
Similarly, we define $\Errg{\tau}:=\bbE[\Df{g}{\tau}]-\Df{g}{\tau}$.

Proposition~\ref{prop: min-B} gives guarantee for the minimization step \cite{moothedath2022fast}.
\begin{proposition}\label{prop: min-B}
    (Min step for $\mat{B}$, Theorem~5.2 \cite{moothedath2022fast}) Assume that, for all $g\in[L]$, $\SD{\U{g}{\tau-1}}{\Ustar}\leqslant\delg{\tau-1}$, where $\delg{\tau-1}\geqslant 0$.
    If $n\geqslant\max(\log T,\log d, r)$, then, with probability at least $1-\exp(\log T+r-cn)$,
    \begin{enumerate}
        \item $\|\bt{\tau}\|\leqslant1.1\|\mat{b}_t^\star\|$
        \item $\|\ttat{\tau}-\tta_t^\star\|\leqslant1.4\delg{\tau-1}\|\mat{b}_t^\star\|$
        \item $\|\Ttat{\tau}-\Ttastar\|_F\leqslant1.4\sqrt{r}\delg{\tau-1}\sigmax$.
    \end{enumerate}
    If $\delg{\tau-1}\leqslant0.02/\sqrt{r}\kappa^2$ and $\max_{g\neq g'}\|\U{g}{\tau-1}-\U{g'}{\tau-1}\|_F\leqslant \rhog{\tau-1}$ with $\rhog{\tau-1}\leqslant c/\sqrt{r}\kappa^2$, then the above implies that
    \begin{enumerate}[label=\alph*)]
        \item $\minsval{\B{}{\tau}}\geqslant 0.9\sigmin$
        \item $\maxsval{\B{}{\tau}}\leqslant1.1\sigmax$.\hfill$\Box$
    \end{enumerate}
\end{proposition}

Proposition~\ref{prop: Gradient deviation} incorporates Lemmas 5.6 and 5.7 of \cite{moothedath2022fast}, which provides the expectation bounds of (partial) sum of local gradients and its deviation around the expectation.
\begin{proposition}\label{prop: Gradient deviation}
(Gradient deviation, Lemmas 5.6 and 5.7 \cite{moothedath2022fast})
Assume that, for all $g\in[L]$, $\SD{\U{g}{\tau-1}}{\Ustar}\leqslant\delg{\tau-1}$ and $\|\UconsErr{g,g'}{\tau-1}\|_F\leqslant\rhog{\tau-1}$. Also, $nT\geqslant C\kappa^4\mu^2dr$ and $n\geqslant \max(\log T,\log d, r)$.
Then, the following hold:
    \begin{enumerate}
        \item $\bbE[\gradU{\tau}]=n(\Ttat{\tau}-\Ttastar){\B{}{\tau}}^\top$
        \item $\|\bbE[\gradU{\tau}]\|\leqslant 1.54n\delg{\tau-1}\sigmax^2$
        \item If $\delg{\tau-1}<c/\sqrt{r}\kappa^2$ and $\rhog{\tau-1}\leqslant c/\sqrt{r}\kappa^2$, then with probability at least $1-\exp\lr{C(d+r)-c\frac{\epsilon_1^2nT}{\kappa^4\mu^2r}}-\exp(\log T+r-cn)$,
        $$\|\Err{\tau}\|\leqslant\epsilon_1\delg{\tau-1}n\sigmin^2.$$
       \item $\bbE[\sum_{g'\neq g}\Df{g'}{\tau}]=n(\Ttasg{\tau}-\Tta_{\setminus g}^\star){\B{\setminus g}{\tau}}^\top$
       \item $\|\bbE[\sum_{g'\neq g}\Df{g'}{\tau}]\|\leqslant1.54 n \delg{\tau-1}\sigmax^2$
       \item Under the same condition and with the same probability of part 3),
       
       $\|\Errg{\tau}\|\leqslant\epsilon_1\delg{\tau-1}n\sigmin^2.$ \hfill$\Box$
   \end{enumerate}
\end{proposition}

For the consensus error analysis, we use the following result from \cite{moothedath2022fast}, which is a matrix version of Proposition \ref{prop: avgcons}. 
\begin{proposition}(Lemma 5.4 \cite{moothedath2022fast})\label{prop: ConsErr_prop}
    If $\Tcongd\geqslant \frac{C}{\log(1/\gamma(\W))}\log(L/\econ)$, then
    \begin{align*}%\label{eqn: ConsErr_prop}
            \|\ConsErr{g}{\tau}\|_F\leqslant\econ\|\InpErr{g}{\tau}\|_F, 
    \end{align*}
    where $\InpErr{g}{\tau}:=\Ubreve{g}{\tau}-\Uhat{\tau}$ is the gap between the input of node $g$ to \textsc{Agree} and the true average.\hfill$\Box$

    We perform QR decomposition to obtain $\U{g}{\tau}$ from $\Utilde{g}{\tau}$.
    The following result is used to analyze the effect of this projection step.
    \begin{proposition}\label{prop: Perteurbed QR}
    (Perturbed QR decomposition \cite{stewart1993perturbation}, \cite{sun1995perturbation})
    Let $\bZtilde_1 \qreq \bZ_1\bR_1$ and $\bZtilde_2 \qreq \bZ_2\bR_2$. Then,
    \begin{align}\label{eqn: Perturbed QR-Q}
        \|\bZ_2-\bZ_1\|_F &\leqslant\sqrt{2}\frac{\|\bZtilde_2-\bZtilde_1\|_F}{\minsval{\bZtilde_1}}\mbox{~and}\\
        \frac{\|\bR_2-\bR_1\|_F}{\|\bR_1\|}&\leqslant\sqrt{2}\frac{\|\bR_1\|}{\|\bR_1^{-1}\|}\frac{\|
        \mat{P}_{\bZtilde_1}(\bZtilde_1-\bZtilde_2)\|_F}{\|\bZtilde_1\|},\label{eqn: Perturbed QR-R}
    \end{align}
    where $\mat{P}_{\bZtilde_1}=\bZtilde_1(\bZtilde_1^\top\bZtilde_1)^{-1}\bZtilde_1^\top$.\hfill$\Box$
\end{proposition}
\end{proposition}
\section{Proof of Theorem \ref{thm: main}}\label{sec: proof of main thm}
In this section, we provide the proof of Theorem \ref{thm: main}.
Recall that Theorem \ref{thm: main} guarantees $\epsilon$-accurate recovery of task parameters stated in part 1) and subspace distance stated in part 2), given that the iteration and sample requirements stated in parts a)-c) hold.
The proof proceeds following four main steps: in the first two steps, we prove parts 1) and 2) of Theorem \ref{thm: main} specifying the required conditions, followed by next two steps explaining how the conditions in parts a)-c) are satisfied.

    \textbf{\em Step 1. Task parameter recovery:} 
    Consider the subspace distance bound at $\tau=\Tgd$, i.e., $\SD{\U{g}{\Tgd}}{\Ustar}\leqslant\delg{\Tgd}$. If 
    \begin{align}
        &\delg{\Tgd}\leqslant\efin\quad \mathrm{and}\label{eqn: cond - efin}\\
        &n\gtrsim \max\lr{\log d,\log T, r},\nn %\label{eqn: cond - sample n}
    \end{align}
     then part 1) of Theorem \ref{thm: main} directly follows by part 2) of Proposition~\ref{prop: min-B} since $\|\tta_t^\star\|^2=\|\mat{b}_t^\star\|^2$.

    \textbf{\em Step 2. Subspace distance decay:}
    We prove that the subspace distance decays up to $\efin$, by establishing the relation of subspace distance at time $\tau$ with respect to $\tau-1$.
    In this and the following steps, we assume that for all $g\in[L]$,
    \begin{itemize}
        \item $\SD{\U{g}{\tau-1}}{\Ustar})\leqslant\delg{\tau-1}$
        \item $\max_{g,g'}\|\UconsErr{g,g'}{\tau-1}\|_F\leqslant\rhog{\tau-1}$
        \item $\max_{g,g'}\|\Pperp\UconsErr{g,g'}{\tau-1}\|_F\leqslant\psig{\tau-1}$.
    \end{itemize}
   Lemma \ref{lem: Subspace distance} characterizes the error decay of the subspace distance.
    We give the proof outline of Lemma \ref{lem: Subspace distance} below. The complete proof is presented in Section \ref{ssc: proof of lemma 1}.
    \begin{lemma}\label{lem: Subspace distance} Let $\eta=\ceta/n\sigmax^2$. Consider the $\tau$-th instance of Algorithm \ref{alg:dec-altgd}. If 
        \begin{align}
            &\delg{\tau-1}\leqslant\frac{0.02}{\sqrt{r}\kappa^2}\label{eqn: cond - delg - lem}\\
            &\rhog{\tau-1}\leqslant\frac{0.1}{\sqrt{r}\kappa^2}=:\cucon\label{eqn: cond - rhog - lem}\\
            &\psig{\tau-1}\leqslant\frac{0.1}{1.21\kappa^2}\delg{\tau-1}=:\cpucon\delg{\tau-1}\label{eqn: cond - psig - lem}\\
            &\|\ConsErr{g}{\tau}\|\leqslant0.01\frac{\ceta}{\kappa^2}=:\ccon\label{eqn: cond - ConsErr - lem}\\
            &\|\Pperp\ConsErr{g}{\tau}\|\leqslant0.01\frac{\ceta}{\kappa^2}\delg{\tau-1}=:\cpcon\delg{\tau-1}\label{eqn: cond - PconsErr - lem},
        \end{align}
        %and if 
        \begin{align}\label{eqn: cond - samples - lem}
            nT\geqslant C\kappa^4\mu^2dr\quad\mathrm{and}\quad n\gtrsim \max(\log T, \log d, r),
        \end{align}
        then, w.h.p, for all $g\in[L]$,
        \begin{align}\label{eqn: SD - result - lemma}
            \SD{\U{g}{\tau}}{\Ustar}\leqslant(1-0.3\ceta/\kappa^2)\delg{\tau-1}=:\delg{\tau}.
        \end{align}  
        % \hfill$\Box$
    \end{lemma}
    \textbf{\em Proof outline of Lemma \ref{lem: Subspace distance}.}
    For complete proof, see Section~\ref{sec: proof of lemmas}. Recall the projection step of the algorithm (Line~\ref{line:proj})
    \begin{align*}
        \Utilde{g}{\tau}&\qreq\U{g}{\tau}\R{g}{\tau}.
    \end{align*}
    Since $\U{g}{\tau}=\Utilde{g}{\tau}(\R{g}{\tau})^{-1}$ and since $\|(\R{g}{\tau})^{-1}\|=1/\minsval{\R{g}{\tau}}=1/\minsval{\Utilde{g}{\tau}}$, we obtain the SD bound as 
    \begin{align}\label{eqn: SD - full - sketch}
        \SD{\U{g}{\tau}}{\Ustar}=\|\Pperp \U{g}{\tau}\|\leqslant\frac{\|\Pperp \Utilde{g}{\tau}\|}{\minsval{\Utilde{g}{\tau}}}.
    \end{align}
    The numerator can be upper bounded by
    \begin{align}
        &\|\Pperp\Utilde{g}{\tau}\|
        \leqslant\|\mat{I}-\eta n\B{}{\tau}{\B{}{\tau}}^\top\|\delg{\tau-1}+\psig{\tau-1}\cdot\|\B{}{\tau}\|^2\nonumber\\
            &\quad+\eta\|\Err{\tau}\| +\|\Pperp\ConsErr{g}{\tau}\|.\label{eqn: lem1 - num - sketch}
    \end{align}
    Also, we derive the lower bound for the denominator as
    \begin{align}
        \minsval{\Utilde{g}{\tau}}&\geqslant1-\rhog{\tau-1}
            -\eta \|\Err{\tau}\|\nonumber\\
            &-\eta \|\bbE[\gradU{\tau}]\|
            -\|\ConsErr{g}{\tau}\|_F.\label{eqn: SD - den - sketch}
    \end{align}
    We use Proposition \ref{prop: min-B} to obtain bounds of $\|\mat{I}-\eta n\B{}{\tau}{\B{}{\tau}}^\top\|$ and $\|\B{}{\tau}\|^2$, and Proposition \ref{prop: Gradient deviation} to obtain the bounds of $\|\Err{\tau}\|$ and $\|\bbE[\gradU{\tau}]\|$.
    Note that using these propositions requires subspace estimates to satisfy Eqs.~\eqref{eqn: cond - delg - lem} and \eqref{eqn: cond - rhog - lem} and samples at least as many as stated in Eq.~\eqref{eqn: cond - samples - lem}. The results are probabilistic.
    Given that the conditions in Eqs.~\eqref{eqn: cond - rhog - lem}-\eqref{eqn: cond - PconsErr - lem} hold, plugging the bounds in Eqs.~\eqref{eqn: lem1 - num - sketch} and \eqref{eqn: SD - den - sketch} for Eq.~\eqref{eqn: SD - full - sketch} yields
    \begin{align*}
        \|\Pperp \U{g}{\tau}\|
        &\leqslant\frac{\|\Pperp \Utilde{g}{\tau}\|}{\minsval{\Utilde{g}{\tau}}}\nonumber\\
        &\leqslant\frac{(1-0.6\ceta/\kappa^2)\delg{\tau-1}}{1-0.15\ceta/\kappa^2}\nonumber\\
        &\leqslant(1-0.6\ceta/\kappa^2)\delg{\tau-1}(1+0.3\ceta/\kappa^2)\nonumber\\
        &\leqslant(1-0.3\ceta/\kappa^2)\delg{\tau-1}
        =:\delg{\tau}
    \end{align*}
    where we use $(1-x)^{-1}\leqslant(1+2x)$ for $|x|\leqslant1/2$.\qed
    
    The proof outline of Lemma \ref{lem: Subspace distance} explains how the subspace distance decay is guaranteed with high probability and the reason why the conditions in Eqs.~\eqref{eqn: cond - delg - lem}-\eqref{eqn: cond - samples - lem} are needed.
    In the next step, we analyze how these requirements can be met.

\textbf{\em Step 3. Iteration and sample complexities of GD step:} 
\textbf{\em Step 3.1. Bounding consensus error, inter-node consensus error and their projections:} 
    The following lemma establishes the connection between the number of consensus iterations $\Tcongd$ and the consensus error $\|\ConsErr{g}{\tau}\|$, and its projection $\|\Pperp\ConsErr{g}{\tau}\|$.
    The proof is in Section \ref{ssc: proof of lemma 2}. 
    \begin{lemma}\label{lem: Conserr}
    (Bounding consensus error) 
    Suppose the conditions given in Eqs.~\eqref{eqn: cond - delg - lem}-\eqref{eqn: cond - psig - lem} and Eq.~\eqref{eqn: cond - samples - lem} hold. If
    \begin{align*}
        \Tcongd\geqslant C\frac{1}{\log(1/\gamma(\W))}\log(L/\econ),
    \end{align*}
    then, w.h.p, we have the following bounds of consensus errors and their projections onto complement of $\Ustar$. 
    \begin{align}\label{eqn: ConsErr - bound}
            \|\ConsErr{g}{\tau}\|_F
            &{\leqslant\econ\Ccon} \mbox{~  and}
    \end{align}
    \begin{align}\label{eqn: PConsErr - bound}
        \|\Pperp\ConsErr{g}{\tau}\|_F
        &{\leqslant\econ\Cpcon\delg{\tau-1}},
    \end{align}
    {for all $g\in[L]$, where $\Ccon:=\sqrt{r}\frac{0.1\ceta+0.034\ceta L}{\kappa^2}$
     and $\Cpcon:=\sqrt{r}(\frac{0.1}{1.21\kappa^2}+1.4\ceta L)$.}\hfill$\Box$

\end{lemma}

Next, the following lemma analyzes the change that the bounds of inter-node consensus error and its projection undergo during update.
The proof is presented in Section \ref{ssc: proof of lemma 3}.
\begin{lemma}\label{lem: Bounding UconsErr} (Bounding inter-node consensus error)
    Suppose conditions Eqs.~\eqref{eqn: cond - delg - lem}-\eqref{eqn: cond - ConsErr - lem} and Eq.~\eqref{eqn: cond - samples - lem} holds. If
    \begin{align*}
        \Tcongd\geqslant C\frac{1}{\log(1/\gamma(\W))}\log(L/\econ),
    \end{align*}
    then, w.h.p, we have the following bounds of inter-node consensus errors and their projections onto complement of $\Ustar$. 
    \begin{align}\label{eqn: uconserr - bound}
        \|\UconsErr{g,g'}{\tau}\|_F
        % \leqslant\econ\sqrt{r}(3.1\rhog{\tau-1}+5.1\ceta L\sqrt{r}\delg{\tau-1})\\
        &{\leqslant\econ\Cucon}=:\rhog{\tau}\mbox{~and}
    \end{align}
    \begin{align}\label{eqn: Puconserr - bound}
        % &\|\Pperp\UconsErr{g,g'}{\tau}\|\\
        % &\leqslant\econ\sqrt{r}(3.1\psig{\tau-1}+(4\rhog{\tau-1}+3.3\ceta L)\delg{\tau-1})\\
        &{\|\Pperp\UconsErr{g,g'}{\tau}\|_F\leqslant\econ\Cpucon\delg{\tau-1}}=:\psig{\tau}
    \end{align}
    {for all $g,g'\in[L]$, where $\Cucon:=\sqrt{r}\frac{0.31\ceta+0.11\ceta L}{\kappa^2}$ and $\Cpucon=\sqrt{r}(\frac{0.31}{1.21\kappa^2}+0.4\frac{\ceta}{\kappa^2}+3.3\ceta L)$ }.\hfill$\Box$

\end{lemma}
\begin{remark} (Bounding inter-node consensus error)\label{remark: inter-node controllability}
    Our analysis relies on node-wise consistency conditions in Eqs.~\eqref{eqn: cond - rhog - lem} and \eqref{eqn: cond - psig - lem}, i.e., small node-wise consensus error.
    Lemma \ref{lem: Bounding UconsErr} shows that the inter-node consensus error scales proportionally with $\econ$, and hence can be reduced by increasing the number of consensus iterations $\Tcongd$.
    This is because our algorithm performs consensus directly on the local update $\Ubreve{g}{\tau}$, producing close enough averaged update $\Utilde{g}{\tau}$ at each iteration.
    Consequently, even if the inter-node disagreement grows during the Dif-AltGDmin steps, running enough agreement rounds ensures that local estimates remain close.
    In contrast, the earlier method in \cite{moothedath2022fast} applies agreement only to gradients. 
    Since gradients do not contain the nodes' estimates, the resulting inter-node consensus error follows the recursion
    \begin{align}\label{eqn: rho - earlier}
        \rhog{\tau}=1.7(\rhog{\tau-1}+2\eta \|\ConsErr{g}{\tau}\|_F),
    \end{align} 
    where $\rhog{\tau}:= \max _{g,g'}\|\UconsErr{g,g'}{\tau}\|_F$ and $\|\ConsErr{g}{\tau}\|_F\leqslant \econ\epsilon_1 n\sigmin^2$.
    Observe that only the second term is proportional to $\econ$, while the factor $1.7>1$ makes the recursion inherently expansive. Hence, once the inter-node error grows, additional agreement iterations cannot reverse its increase.
\end{remark}

\textbf{\em Step 3.2. Iteration complexities at Dif-AltGDmin step:}    
    Suppose the conditions Eqs.~\eqref{eqn: cond - delg - lem}-\eqref{eqn: cond - samples - lem} are satisfied for all $\tau=1,...,\Tgd$. 
    Given $\delg{0}$, Lemma \ref{lem: Subspace distance} guarantees that repeating $\Tgd$ rounds of Dif-AltGDmin gives
    \begin{align*}
        \delg{\Tgd} = \big(1-0.3\frac{\ceta}{\kappa^2}\big)^{\Tgd}\delg{0}.
    \end{align*}
    Based on this, we obtain the expression for $\Tgd$ that ensures $\delg{\Tgd}\leqslant\efin$ (Eq.~\eqref{eqn: cond - efin}).
    Taking logarithms and rearranging expression for $\Tgd$, we obtain
    \begin{align*}
        &\log \delg{\Tgd} 
        = \Tgd \log\Big(1-0.3\frac{\ceta}{\kappa^2}\Big) + \log \delg{0} \leqslant \log \efin, \nonumber \\
        &\implies \Tgd \geqslant -\frac{\log(\delg{0}/\efin)}{\log\big(1-0.3\ceta/\kappa^2\big)}.
    \end{align*}
    If $\delg{0}\leqslant\frac{c}{\sqrt{r}\kappa^2}$, we can obtain the bound 
    \begin{align*}
        \Tgd \geqslant C \frac{\kappa^2}{\ceta} \log(1/\efin)
    \end{align*}
    by using the inequality 
    $
    -\log(1-x) = \int_0^x \frac{1}{1-t} dt \leqslant \frac{x}{1-x} \leqslant 2x$ for $0\leqslant x \leqslant \frac{1}{2},
    $
    and absorbing $\log \delg{0}\leqslant \log (\frac{c}{\sqrt{r}\kappa^2})\ll \log (1/\efin)$ into a constant $C$.
    This establishes the required number of Dif-AltGDmin steps $\Tgd$ to guarantee Eq.~\eqref{eqn: cond - efin}.

    Next, we obtain the number of consensus iterations $\Tcongd$.
    Previous steps state that our analysis remains valid if the conditions Eqs.~\eqref{eqn: cond - rhog - lem}-\eqref{eqn: cond - PconsErr - lem} hold at each $\tau=1,...,\Tgd$. (Also, Eq.~\eqref{eqn: cond - delg - lem} should hold.)
    { From Step 3.1, we know that (inter-node) consensus errors and their projections are bounded as in Eqs.~\eqref{eqn: ConsErr - bound}-\eqref{eqn: Puconserr - bound}, if the conditions in Eqs.~\eqref{eqn: cond - rhog - lem}-\eqref{eqn: cond - PconsErr - lem} hold. 
    From Eqs.~\eqref{eqn: cond - ConsErr - lem} and \eqref{eqn: ConsErr - bound}, if we set $\econ\leqslant\frac{\ccon}{\Ccon}$ in round $\tau$, then Eq.~\eqref{eqn: cond - ConsErr - lem} holds in round $\tau$.
    Similarly, from Eqs.~\eqref{eqn: cond - PconsErr - lem} and \eqref{eqn: PConsErr - bound}, if we set $\econ\leqslant\frac{\cpcon}{\Cpcon}$ in round $\tau$, then Eq.~\eqref{eqn: cond - PconsErr - lem} holds in round $\tau$.
    Further, from Eqs.~\eqref{eqn: cond - rhog - lem} and \eqref{eqn: uconserr - bound}, if we set $\econ\leqslant\frac{\cucon}{\Cucon}$ in round $\tau$, then Eq.~\eqref{eqn: cond - rhog - lem} holds in round $\tau+1$. 
    Similarly, from Eqs.~\eqref{eqn: cond - psig - lem}, \eqref{eqn: Puconserr - bound}, and using~\eqref{eqn: SD - result - lemma}, if we set $\econ\leqslant\frac{\cpucon}{\Cpucon}(1-0.3\frac{\ceta}{\kappa^2})$ in round $\tau$, then Eq.~\eqref{eqn: cond - psig - lem} holds in round $\tau+1$.
    }
    Since these must hold simultaneously, we deduce that the consensus accuracy $\econ$ should satisfy
    \begin{align}\label{eqn: econ - min}
        \scalemath{0.9}{\econ\leqslant\min\lr{\frac{\ccon}{\Ccon}, \frac{\cpcon}{\Cpcon},\frac{\cucon}{\Cucon}, \frac{\cpucon(1-0.3\frac{\ceta}{\kappa^2})}{\Cpucon}}}.
    \end{align}
    
    Both Lemmas \ref{lem: Conserr} and \ref{lem: Bounding UconsErr} state that such an accuracy $\econ$ can be guaranteed by at least 
    \begin{align}\label{eqn: Tcongd - min}
        \Tcongd&\geqslant C\frac{1}{\log(1/\gamma(\W))}\log(L/\econ)\nonumber\\
        &= C\frac{1}{\log(1/\gamma(\W))}(\log L+\log(1/\econ))
    \end{align}
    %whenever the AvgCons algorithm runs.
   consensus iterations.
    Rewriting Eq.~\eqref{eqn: econ - min} as
    \begin{align*}
        \scalemath{0.9}{\log\lr{\frac{1}{\econ}}
        \geqslant
        \log \max\lr{\frac{\Ccon}{\ccon}, \frac{\Cpcon}{\cpcon},\frac{\Cucon}{\cucon}, \frac{\Cpucon}{\cpucon(1-0.3\frac{\ceta}{\kappa^2})}}}
    \end{align*}
    and by substituting the explicit expressions of these constants, we require $\log(1/\econ)$ to be greater than all of the following
    \begin{itemize}
        \item $\log(\frac{\Ccon}{\ccon})
            \geqslant\log(\frac{\sqrt{r}(0.1\ceta/\kappa^2+0.034\ceta L/\kappa^2)}{\frac{0.01\ceta}{\kappa^2}})$;
        \item $\log(\frac{\Cpcon}{\cpcon})
            \geqslant\log(\frac{\sqrt{r}(0.1/1.21\kappa^2+1.4\ceta L)}{\frac{0.01\ceta}{\kappa^2}})$;
        \item $\log(\frac{\Cucon}{\cucon})
            \geqslant\log(\frac{\sqrt{r}(0.31\ceta/\kappa^2+5.1\ceta L\sqrt{r}\cdot0.02/\sqrt{r}\kappa^2}{0.1\ceta/\kappa^2})$;
        \item ${\log\lr{\frac{\Cpucon}{\cpucon(1-0.3\frac{\ceta}{\kappa^2})}}
            \geqslant\log\lr{\frac{\sqrt{r}(\frac{0.31}{1.21\kappa^2}+0.4\frac{\ceta}{\kappa^2}+3.3\ceta L)}{\frac{0.1}{1.21\kappa^2}(1-0.3\ceta/\kappa^2)}}}$.
    \end{itemize}
    Thus, we require
    \begin{align}\label{eqn: econ - order}
        \log(1/\econ)\gtrsim\log L+\log r+\log\kappa.
    \end{align}   
    Combining Eqs.~\eqref{eqn: Tcongd - min}, \eqref{eqn: econ - order}, thecommunication complexity
       \begin{align}
        \Tcongd\geqslant C\frac{1}{\log(1/\gamma(\W))}(\log L+\log r +\log\kappa)\label{eqn: Tcongd}.
    \end{align}
    Note that setting $\Tcongd$ in round $\tau$ guarantees Eqs.~\eqref{eqn: cond - rhog - lem}, \eqref{eqn: cond - psig - lem} hold in $\tau+1$. 
    To ensure these conditions hold for all rounds, especially at $\tau=1$, we require proper initialization of  $\delg{0}$, $\rhog{0}$ and $\psig{0}$. This initialization is achieved by choosing $\Tpm$ and $\Tconinit$ as specified in Step 4.

   \begin{remark}\emph{How Dif-AltGDmin improve communication complexity as compared to Dec-AltGDmin?}
        While prior approach in \cite{moothedath2022fast} requires of the order of
        $C\frac{1}{\log(1/\gamma({\W}))}(\log L + \Tgd+\log(1/\efin))$ communication rounds per iteration,
        our method only requires the communication rounds given in Eq.~\eqref{eqn: Tcongd}, where $\log r$, $\log \kappa$, $\log L$ are significantly smaller than $\Tgd$ or $\log(1/\efin)$.
        This improvement is achieved by incorporating \emph{projected} consensus errors $\|\Pperp\UconsErr{g,g'}{\tau}\|_F$ and $\|\Pperp\ConsErr{g}{\tau}\|_F$ into the analysis of Eq.~\eqref{eqn: lem1 - num - sketch}. In particular, observe that the required bounds for the guarantees to hold and the bounds we achieve for these consensus error parameters, both scale proportionally with $\delg{\tau-1}$ (see Eqs.~\eqref{eqn: cond - psig - lem} and \eqref{eqn: Puconserr - bound}, and 
        Eqs.~\eqref{eqn: cond - PconsErr - lem} and \eqref{eqn: PConsErr - bound}). 
        Hence, the dependency of $\econ$ on $\delg{\tau-1}$ is eliminated, as shown in Eq.~\eqref{eqn: econ - min}.
        In contrast, \cite{moothedath2022fast} obtains bound for $\|\Pperp\Utilde{g}{\tau}\|$ based on \emph{un-projected} (inter-node) consensus errors $\|\UconsErr{g,g'}{\tau-1}\|$ and $\|\ConsErr{g}{\tau}\|$, both required to scale with $\delg{\tau-1}$ (see Eq. (6) of \cite{moothedath2022fast} {and below}).
        However, as noted in Remark \ref{remark: inter-node controllability} and Eq.~\eqref{eqn: rho - earlier}, $\rhog{\tau}$ only accumulates and $\|\ConsErr{g}{\tau}\|_F$ does not scale with $\delg{\tau-1}$.
        Since achieving $\efin$-accurate recovery requires these quantities to remain below $\efin$ even after $\Tgd$ gradient steps, this requires us to choose $\rhog{0}$ and $\econ$ far smaller than $\efin$, resulting in a substantially higher communication cost.

    \end{remark}

    \textbf{\em Step 4: Iteration complexities of initialization step and total sample complexity:}
    {From part 1) and 2) of Proposition~\ref{prop: initialization}, we have $\psig{0}= \rhog{0}=0$, so it suffices to ensure $\delg{0}\leqslant\frac{0.02}{\sqrt{r}\kappa^2}$.
    }

    By Proposition \ref{prop: initialization}, it determines the requirements for the initialization iterations:
    \begin{align}
        &\Tpm\geqslant C\kappa^2\log(d/\delg{0})=C\kappa^2(\log d+\log \kappa)\quad\mathrm{and}\nn\\
        &\Tconinit\geqslant C\frac{1}{\log1/\gamma(\W)}\bigg(\log Ld\kappa+\log \left({1}/{\delg{0}}\right)\bigg)\nn\\
        &\qquad= C\frac{1}{\log1/\gamma(\W)}(\log L+\log d+\log\kappa+\log r)\label{eqn: Tconinit}.
    \end{align}   
    Next, we bound the total number of samples required for initialization and Dif-AltGDmin phases across all tasks.
    For initialization, we need
    $nT\gtrsim\kappa^8\mu^2(d+T)r^2=:\ninit T$, 
    and for each Dif-AltGDmin iteration,
    we need $n T\gtrsim C\kappa^4\mu^2dr=:\ngd T$.
    Thus the total sample complexity is 
    \begin{align*}
        nT&\gtrsim \ninit T+\Tgd\cdot \ngd T\\
        &=\kappa^8\mu^2(d+T)r^2+C\kappa^6\mu^2dr\log(1/\efin)\\
        &=C\kappa^6\mu^2(d+T)r(\kappa^2r+\log(1/\efin)).
    \end{align*} 
    This completes the proof of Theorem \ref{thm: main}.\qed
    \begin{remark}\label{remark: robustness to connectivity} (\textit{Robustness to weak connectivity})
        { Recall the network connectivity expression in Eq.~\eqref{eqn: connectivity}. Consider $\Tcon :=\max{(\Tcongd,\Tconinit)}=\Tconinit$. 
While Dif-AltGDmin requires communication rounds as in Eq.~\eqref{eqn: Tconinit} %$\log(1/\econ)\gtrsim \log(Ldr\kappa)$
to achieve $\efin$-accurate estimate,
Dec-AltGDmin requires a significantly tighter consensus accuracy $\log(1/\econ)\gtrsim \log (Ld\kappa\lr{1/\efin}^{\kappa^2})$
(Theorem 4.1 in \cite{moothedath2022fast}).
Since $\efin$ is typically very small, the consensus accuracy term $\log(1/\econ)$ for Dec-AltGDmin becomes substantially larger. 
Substituting $\log(1/\econ)$ expressions into Eq.~\eqref{eqn: connectivity}, we conclude that that Dec-AltGDmin imposes a much tighter upper bound on $\gamma(\W)$.
Smaller $\gamma(\W)$ indicates a denser network, meaning that Dec-AltGDmin requires a stronger network connectivity to guarantee its convergence.
Dif-AltGDmin, on the other hand, is robust to larger $\gamma(\W)$s and hence effective even under sparse networks.
}

    \end{remark}
\section{Proof of Lemmas}\label{sec: proof of lemmas}
\subsection{Proof of Lemma \ref{lem: Subspace distance}}\label{ssc: proof of lemma 1}
   Recall the projected gradient descent step of the algorithm
    \begin{align*}
        \Utilde{g}{\tau}&=\AvgCons(\U{g}{\tau-1}-\eta L \Df{g}{\tau},\graph,\Tcongd)\nonumber\\
        \Utilde{g}{\tau}&\qreq\U{g}{\tau}\R{g}{\tau}.%\label{eqn: update rule}.
    \end{align*}
    Since $\U{g}{\tau}=\Utilde{g}{\tau}(\R{g}{\tau})^{-1}$ and $\|(\R{g}{\tau})^{-1}\|=\frac{1}{\minsval{\R{g}{\tau}}}=\frac{1}{\minsval{\Utilde{g}{\tau}}}$, the subspace distance can be bounded as follows:
    \begin{align}\label{eqn: SD-full expression}
        \SD{\U{g}{\tau}}{\Ustar}=\|\Pperp \U{g}{\tau}\|\leqslant\frac{\|\Pperp \Utilde{g}{\tau}\|}{\minsval{\Utilde{g}{\tau}}}.
    \end{align}
    We obtain the upper bound of the numerator and lower bound of the denominator separately.
    
    First, consider the numerator.
    By adding and subtracting $\Pperp\Uhat{\tau}$ to $\Pperp\Utilde{g}{\tau}$, we obtain
    \begin{align}\label{eqn: SD-perturb}
        \Pperp\Utilde{g}{\tau}&=\Pperp\Uhat{\tau}+(\Pperp\Utilde{g}{\tau}-\Pperp\Uhat{\tau})\nonumber\\
        &=\Pperp\Uhat{\tau}+\Pperp\ConsErr{g}{\tau}.
    \end{align}
    Adding and subtracting $\eta \bbE[\gradU{\tau}]$ to $\Pperp\Uhat{\tau}$ gives
    \begin{align}
        &\Pperp\Uhat{\tau}=\Pperp\Ubar{\tau-1}-\eta \Pperp\gradU{\tau}\nn\\
        &=\Pperp\Ubar{\tau-1}
            +\eta \Pperp(\Err{\tau}-\bbE[\gradU{\tau}])\label{eq:Pperp}.
    \end{align}
    
    We simplify $\Pperp\bbE[\gradU{\tau}]$ using part 1) of Proposition \ref{prop: Gradient deviation}, the fact that $\Pperp\Ustar=0$ and
    $\sum_{g\in[L]}\sum_{t\in\Sg}\bt{\tau}\bt{\tau}^\top=\sum_{g\in[L]}\B{g}{\tau}{\B{g}{\tau}}^\top=\B{}{\tau}{\B{}{\tau}}^\top$,
    \begin{align*}
        &\Pperp\bbE[\gradU{\tau}]
        =n\Pperp(\Ttat{\tau}-\Ttastar){\B{}{\tau}}^\top\nn\\
        &=n\Pperp\sum_{g=1}^L\sum_{t\in\mathcal{S}_g}(\ttat{\tau}-\tta_t^\star){\bt{\tau}}^\top\nn\\
        &=n\Pperp\sum_{g=1}^L\sum_{t\in\mathcal{S}_g}(\U{g}{\tau-1}\bt{\tau}-\Ustar\mat{b}_t^\star){\bt{\tau}}^\top\nn\\
        &=n\Pperp\sum _{g=1}^L\U{g}{\tau-1}\B{g}{\tau}{\B{g}{\tau}}^\top.
    \end{align*}
    By adding and subtracting $\Ubar{\tau-1}$ to $\U{g}{\tau-1}$,
    \begin{align}
        &\Pperp\bbE[\gradU{\tau}]\nn\\
        &=n\Pperp\sum _{g=1}^L(\U{g}{\tau-1}-\Ubar{\tau-1}+\Ubar{\tau-1})\B{g}{\tau}{\B{g}{\tau}}^\top\nn\\
        % &=n\Pperp\sum _{g=1}^L\Ubar{\tau-1}\B{g}{\tau}{\B{g}{\tau}}^\top \\
            % &\quad+\eta n\Pperp\sum _{g=1}^L(\Ubar{\tau-1}-\U{g}{\tau-1})\B{g}{\tau}{\B{g}{\tau}}^\top.\\
        &=n\Pperp\Ubar{\tau-1}\B{}{\tau}{\B{}{\tau}}^\top\nn\\
            &\quad- n\Pperp\sum _{g=1}^L(\Ubar{\tau-1}-\U{g}{\tau-1})\B{g}{\tau}{\B{g}{\tau}}^\top.\label{eq:Pperp_2}
    \end{align}
    Using Eqs.~\eqref{eq:Pperp} and \eqref{eq:Pperp_2}, we rewrite Eq.~\eqref{eqn: SD-perturb} as
    \begin{align*}
        \Pperp\Utilde{g}{\tau}&=\Pperp\Ubar{\tau-1}(I-\eta n \B{}{\tau}{\B{}{\tau}}^\top)\\
            &\quad+\eta n\Pperp\sum _{g=1}^L(\Ubar{\tau-1}-\U{g}{\tau-1})\B{g}{\tau}{\B{g}{\tau}}^\top\\
            &\quad+\eta\Pperp\Err{\tau} +\Pperp\ConsErr{g}{\tau}.
    \end{align*}
    Applying norm, triangular and Cauchy-Schwarz inequalities,
    \begin{align}
        &\|\Pperp\Utilde{g}{\tau}\|\leqslant\|\Pperp\Ubar{\tau-1}(I-\eta n \B{}{\tau}{\B{}{\tau}}^\top)\|\nn\\
            &\quad+\eta n\|\Pperp\sum _{g=1}^L(\Ubar{\tau-1}-\U{g}{\tau-1})\B{g}{\tau}{\B{g}{\tau}}^\top\|\nn\\
            &\quad+\eta\|\Pperp\Err{\tau}\| +\|\Pperp\ConsErr{g}{\tau}\|\nn\\
        &\leqslant\|\Pperp\Ubar{\tau-1}\|\|(I-\eta n \B{}{\tau}{\B{}{\tau}}^\top)\|\nn\\
            &\quad+\eta n\|\Pperp\sum _{g=1}^L(\Ubar{\tau-1}-\U{g}{\tau-1})\|\cdot\max_g \|\B{g}{\tau}\|^2\nn\\
            &\quad+\eta\|\Pperp\|\|\Err{\tau}\| +\|\Pperp\ConsErr{g}{\tau}\|.\label{eq:Pperp_3}
    \end{align}
    Consider the first term.
    Since we have assumed $\SD{\U{g}{\tau-1}}{\Ustar}=\|\Pperp\U{g}{\tau-1}\|\leqslant\delg{\tau-1}$ for all $g\in[L]$, it follows that
    \begin{align*}
        &\|\Pperp\Ubar{\tau-1}\|=\|\Pperp\frac{1}{L}\sum_{g=1}^L\U{g}{\tau-1}\|\\
        &\leqslant\frac{1}{L}\sum_{g=1}^L\|\Pperp\U{g}{\tau-1}\|\leqslant\frac{1}{L}\sum_{g=1}^L\delg{\tau-1}=\delg{\tau-1}.
    \end{align*}
    Also, Proposition \ref{prop: min-B} says that if $\delg{\tau-1}\leqslant0.02/\sqrt{r}\kappa^2$ and $\rhog{\tau-1}\leqslant0.1\ceta/\kappa^2$,
    \begin{align*}
        \mineval{\mat{I}-\eta n \B{}{\tau}{\B{}{\tau}}^\top}=1-\eta n \|\B{}{\tau}\|^2\geqslant 1-1.21\eta n \sigmax^2.
    \end{align*}
    Thus if $\eta<0.5/n\sigmax^2$, the matrix $\mat{I}-\eta n \B{}{\tau}{\B{}{\tau}}^\top$ is positive semi-definite, so that
    \begin{align*}
        \|I-\eta n \B{}{\tau}{\B{}{\tau}}^\top\|
            &=\maxeval{\mat{I}-\eta n \B{}{\tau}{\B{}{\tau}}^\top}\\
            &\leqslant1-0.81\eta n\sigmin^2.
    \end{align*}
    
    Next, consider the second term.
    \begin{align*}
        &\|\Pperp(\Ubar{\tau-1}-\U{g}{\tau-1})\|\\
        &=\|\frac{1}{L}\sum_{g'=1}^L\Pperp(\U{g'}{\tau-1}-\U{g}{\tau-1})\|_F\\
        &\leqslant\max_g\|\Pperp(\U{g'}{\tau-1}-\U{g}{\tau-1})\|_F\\
        &\leqslant\max_g\|\Pperp\UconsErr{g,g'}{\tau-1}\|_F\leqslant\psig{\tau-1}.
    \end{align*}
    It follows that $\|\B{g}{\tau}\|\leqslant \|\B{}{\tau}\|\leqslant 1.1 \sigmax$ by part b) of Proposition \ref{prop: min-B}.
      
    For the third term, we use part 3) of Proposition \ref{prop: Gradient deviation} with $\epsilon_1=0.1$ and the fact that $\|\Pperp\|=1$ to get
    \begin{align*}
        \|\Pperp\|\|\Err{\tau}\|\leqslant0.1 n\delg{\tau-1}\sigmin^2.
    \end{align*}
    
    By setting $\eta=\ceta/n\sigmax^2$ and assuming that
    \begin{align}
        &\psig{\tau-1}\leqslant\frac{0.1}{1.21\kappa^2}\delg{\tau-1}=:\cpucon\delg{\tau-1}\label{eqn: cond - psig}\\
        &\|\Pperp\ConsErr{g}{\tau}\|\leqslant0.01\frac{\ceta}{\kappa^2}\delg{\tau-1}=:\cpcon\delg{\tau-1}\label{eqn: cond - PconsErr},
    \end{align}
    we can further simplify Eq.~\eqref{eq:Pperp_3}
    \begin{align}
        &\|\Pperp\Utilde{g}{\tau}\|
        \leqslant(1-0.81\eta n\sigmin^2)\delg{\tau-1}
            +0.1\eta n\delg{\tau-1}\sigmin^2\nonumber\\
            &\quad 
            +1.21 \eta n\sigmax^2\cdot \frac{0.1}{1.21\kappa^2}\delg{\tau-1} 
            +0.01 \frac{\ceta}{\kappa^2}\delg{\tau-1}\nonumber\\
        &\leqslant(1-0.6\frac{\ceta}{\kappa^2})\delg{\tau-1}\label{eqn: SD-num result}.
    \end{align}
    This establishes the upper bound of the numerator of Eq.~\eqref{eqn: SD-full expression}.
    
    Next, we obtain the lower bound of the denominator $\minsval{\Utilde{g}{\tau}}$ of Eq.~\eqref{eqn: SD-full expression}.
    Adding and subtracting $\Uhat{\tau}$ to $\Utilde{g}{\tau}$,
    \begin{align*}\label{eqn: SD-den}
        &\minsval{\Utilde{g}{\tau}}
        =\minsval{\Uhat{\tau}+\Utilde{g}{\tau}-\Uhat{\tau}}\nonumber\\
        &=\minsval{\Uhat{\tau}+\ConsErr{g}{\tau}}\nonumber\\
        &\geqslant\minsval{\Uhat{\tau}}
            -\|\ConsErr{g}{\tau}\|_F\nonumber\\
        &=\minsval{\Ubar{\tau-1}
            -\eta \gradU{\tau}}-\|\ConsErr{g}{\tau}\|_F\nonumber\\
        &=\minsval{\Ubar{\tau-1}
            -\eta (\Err{\tau}-\bbE[\gradU{\tau}])}
            -\|\ConsErr{g}{\tau}\|_F\nonumber\\
        &\geqslant\minsval{\Ubar{\tau-1}}
            -\eta (\|\Err{\tau}\|+ \|\bbE[\gradU{\tau}]\|)
            -\|\ConsErr{g}{\tau}\|_F\nonumber
    \end{align*}
    where we use the definitions of $\ConsErr{g}{\tau}$ and $\Uhat{\tau}$, and add and subtract $\bbE[\gradU{\tau}]$ to get the last equality. Weyl's inequality is used for both inequalities.
    
    The lower bound of the first term $\minsval{\Ubar{\tau-1}}$ is derived as follows.
    \begin{align*}
        &\minsval{\Ubar{\tau-1}}=\min_g\minsval{\U{g}{\tau-1}+\Ubar{\tau-1}-\U{g}{\tau-1}}\\
        &\geqslant\min_g\minsval{\U{g}{\tau-1}}-\max_g\|(\frac{1}{L}\sum_{g'=1}^L\U{g'}{\tau-1})-\U{g}{\tau-1}\|\\
        &=\min_g\minsval{\U{g}{\tau-1}}-\max_{g}\|\frac{1}{L}\sum_{g'=1}^L(\U{g'}{\tau-1}-\U{g}{\tau-1})\|\\
        &\geqslant\min_g\minsval{\U{g}{\tau-1}}-\frac{1}{L}\sum_{g'=1}^L\max_{g\neq g'}\|\U{g}{\tau-1}-\U{g'}{\tau-1}\|\\
        &=1-\max_{g\neq g'}\|\UconsErr{g,g'}{\tau-1}\|_F=1-\rhog{\tau-1}.
    \end{align*}
    Given that $\delg{\tau-1}\leqslant\frac{0.02}{\sqrt{r}\kappa^2}$ and $\rhog{\tau-1}\leqslant0.1\frac{\ceta}{\kappa^2}$, the bounds of the second and third terms directly follow by parts 2) and 3) of Proposition \ref{prop: Gradient deviation}. 
    Assuming that 
    \begin{align}
        \|\ConsErr{g}{\tau}\|_F\leqslant0.01\frac{\ceta}{\kappa^2}=:\ccon,
    \end{align} 
    we can simplify the lower bound of $\minsval{\Utilde{g}{\tau}}$ as follows:
    \begin{align}
        &\minsval{\Utilde{g}{\tau}}
        \geqslant1-\rhog{\tau-1}-0.1\eta\delg{\tau-1}n\sigmin^2\nonumber\\
            &-1.54\eta n\sqrt{r}\delg{\tau-1}\sigmax^2
            -\|\ConsErr{g}{\tau}\|_F\nonumber\\
        &\geqslant1-\frac{0.1\ceta}{\kappa^2}-\eta n (0.1\sigmin^2
            +1.54\sqrt{r}\sigmax^2)\delg{\tau-1} -\frac{0.01\ceta}{\kappa^2}\nonumber\\
       &\geqslant1-\frac{0.11\ceta}{\kappa^2}-\eta n (0.1\sqrt{r}\sigmax^2
            +1.54\sqrt{r}\sigmax^2)\delg{\tau-1}\nonumber\\            
        &\geqslant 1-\frac{0.11\ceta}{\kappa^2}
            -\frac{\ceta}{n\sigmax^2}1.64n\sqrt{r}\sigmax^2\frac{0.02}{\sqrt{r}\kappa^2}\nonumber\\
        &= 1-\frac{0.11\ceta}{\kappa^2}-\frac{0.0328\ceta}{\kappa^2}= 1-\frac{0.15\ceta}{\kappa^2},\label{eqn: SD-den result}
    \end{align}
    where we used the fact $\sigmax\geqslant\sigmin$ and $\sqrt{r}\geqslant 1$ for the second inequality.  
    Substituting the results in Eqs.~\eqref{eqn: SD-num result} and~\eqref{eqn: SD-den result} into Eq.~\eqref{eqn: SD-full expression} completes the proof.
    \begin{align*}
        \|\Pperp \U{g}{\tau}\|
        &\leqslant\frac{\|\Pperp \Utilde{g}{\tau}\|}{\minsval{\Utilde{g}{\tau}}}
        \leqslant\frac{(1-0.6\ceta/\kappa^2)\delg{\tau-1}}{1-0.15\ceta/\kappa^2}\nonumber\\
        &\leqslant(1-0.6\ceta/\kappa^2)\delg{\tau-1}(1+0.3\ceta/\kappa^2)\nonumber\\
        &\leqslant(1-0.3\ceta/\kappa^2)\delg{\tau-1}
        =:\delg{\tau}%\label{eqn: SD - decay}
    \end{align*}
    where we use $(1-x)^{-1}\leqslant(1+2x)$ for $|x|\leqslant1/2$.\qed
%\newpage    
\subsection{Proof of Lemma \ref{lem: Conserr}}\label{ssc: proof of lemma 2}
    Lemma~\ref{lem: Conserr} bounds $\|\ConsErr{g}{\tau}\|_F$ and $\|\Pperp\ConsErr{g}{\tau}\|_F$. In order to obtain them, we use Proposition \ref{prop: ConsErr_prop}, which requires bounding $\|\InpErr{g}{\tau}\|$ and $\|\Pperp\InpErr{g}{\tau}\|$.
    Throughout, we assume that $\Tcongd$ satisfies Eq.~\eqref{eqn: Tcongd - min}.
    \subsubsection{Bounding $\|\ConsErr{g}{\tau}\|_F$}
    Consider $\InpErr{g}{\tau}$ given by
    \begin{align*}
        &\InpErr{g}{\tau}=(\U{g}{\tau-1}-\eta L\Df{g}{\tau})-\Uhat{\tau}\\
        &=\U{g}{\tau-1}-\Ubar{\tau-1}+\eta\gradU{\tau}-\eta L\Df{g}{\tau}\\
        % &{\cblue=\U{g}{\tau-1}-\frac{1}{L}\sum_{g'=1}^L\U{g'}{\tau-1}+\eta\sum_{g'=1}^L\Df{g}{\tau}-\eta L\Df{g}{\tau}}\\
        &=\frac{1}{L}\sum_{g'=1}^L(\U{g'}{\tau-1}-\U{g}{\tau-1})+\eta \sum_{g'\neq g}\Df{g'}{\tau}-\eta(L-1)\Df{g}{\tau}.
    \end{align*}
    By defining $\UconsErr{g,g'}{\tau-1}:=\U{g'}{\tau-1}-\U{g}{\tau-1}$, adding and subtracting $\bbE[\Df{g'}{\tau}]=\sum_{t\in\mathcal{S}_{g'}}n(\ttat{\tau}-\tta_t^\star)\bt{\tau}^\top$ $\forall g'\in[L]$,
    and using the definition $\Errgp{\tau}:=\Df{g'}{\tau}-\bbE[\Df{g'}{\tau}]$,
    \begin{align} 
         &\InpErr{g}{\tau}=\frac{1}{L}\sum_{g'=1}^L(\UconsErr{g,g'}{\tau-1})\nn\\
        &\quad+\eta \sum_{g'\neq g}\left(\Df{g'}{\tau}-\bbE[\Df{g'}{\tau}]+\bbE[\Df{g'}{\tau}]\right)\nn\\
        &\quad-\eta(L-1)\left(\Df{g}{\tau}-\bbE[\Df{g}{\tau}]+\bbE[\Df{g}{\tau}]\right)\nn\\
        % &{\cblue=\frac{1}{L}\sum_{g'=1}^L(\UconsErr{g,g'}{\tau-1})
        % +\eta \sum_{g'\neq g}\left(\Errgp{\tau}+\bbE[\Df{g'}{\tau}]\right)}\nn\\
        % &\quad-\eta(L-1)\left(\Errg{\tau}+\bbE[\Df{g}{\tau}]\right)\nn\\
        &=\frac{1}{L}\sum_{g'=1}^L(\UconsErr{g,g'}{\tau-1})
        +\eta \left(\sum_{g'\neq g}\Errgp{\tau}+\bbE[\sum_{g'\neq g}\Df{g'}{\tau}]\right)\nn\\
        &\quad-\eta(L-1)\left(\Errg{\tau}+\bbE[\Df{g}{\tau}]\right)\label{eqn: lem2 - InpErr}.
    \end{align}        
    By taking norm and using triangular inequality, we obtain
    \begin{align*}
        &\|\InpErr{g}{\tau}\|
        % =\|\frac{1}{L}\sum_{g'=1}^L(\UconsErr{g,g'}{\tau-1})
        %     +\eta \sum_{g'\neq g}\Errgp{\tau}}\\
        % &\quad+\eta\bbE[\sum_{g'\neq g}\Df{g'}{\tau}] -\eta(L-1)\left(\Errg{\tau}+\bbE[\Df{g}{\tau}]\right)\| \\
        \leqslant\frac{1}{L}\cdot L\cdot \max_{g'}\|\UconsErr{g,g'}{\tau}\| +\eta \|\sum_{g'\neq g}\Errgp{\tau}\|\\
        &\quad+ \eta \|\bbE[\sum_{g'\neq g}\Df{g'}{\tau}]\|+\eta(L-1) \left(\|\Errg{\tau}\|+\|\bbE[\Df{g}{\tau}]\|\right).
        \end{align*}
        For the bound of the first term, we assume $\max_{g'}\|\UconsErr{g,g'}{\tau-1}\|\leqslant\rhog{\tau-1}$, and use the result of Proposition \ref{prop: Gradient deviation} with $\epsilon_1=0.1$ to bound the last four terms.
        \begin{align*}
         &\|\InpErr{g}{\tau}\|\leqslant\rhog{\tau-1}
            +\eta(0.1\delg{\tau-1}n\sigmin^2+ 1.54n\delg{\tau-1}\sigmax^2)\\
            &\quad+ \eta(L-1)\left(0.1\delg{\tau-1}n\sigmin^2+ 1.54n\delg{\tau-1}\sigmax^2\right)\\
        &\leqslant \rhog{\tau-1}
            +\eta L(0.1\delg{\tau-1}n\sigmax^2 
            + 1.54n\delg{\tau-1}\sigmax^2)\\
        &\leqslant \rhog{\tau-1}
            +\frac{\ceta}{n\sigmax^2} L( 1.64n\delg{\tau-1}\sigmax^2)\\
        &=\rhog{\tau-1}+1.7\ceta L\delg{\tau-1}.
    \end{align*}
    
    Applying $\sqrt{r}$ for the Frobenius norm bound,
    $$\|\InpErr{g}{\tau}\|_F\leqslant\sqrt{r}\left(\rhog{\tau-1}+1.7\ceta L \delg{\tau-1}\right).$$
    Using Proposition \ref{prop: ConsErr_prop}, {if $\rhog{\tau-1}\leqslant0.1\ceta/\kappa^2$ and $\delg{\tau-1}\leqslant0.02/\sqrt{r}\kappa^2$},
    \begin{align}
        &\|\ConsErr{g}{\tau}\|_F\leqslant\econ\sqrt{r}\lr{\rhog{\tau-1}+1.7\ceta L \delg{\tau-1}}\nonumber\\
        &\leqslant \econ\sqrt{r}(0.1\frac{\ceta}{\kappa^2}+1.7\ceta L\frac{0.02}{\sqrt{r}\kappa^2})\leqslant \econ \Ccon,\label{eqn: ConsErr - Bound}
    \end{align}
    where $\Ccon:=\sqrt{r}\frac{0.1\ceta+0.034\ceta L}{\kappa^2}$.

    \subsubsection{Bounding $\|\Pperp\ConsErr{g}{\tau}\|_F$}
    Note that the agreement algorithm is a linear operation as it only does scalar multiplication and addition. Thus, if $\Z{g}{\mathrm{out}}=\AvgCons(\Z{g}{\mathrm{in}},\graph,\Tcongd)$, then
    $\Pperp \Z{g}{\mathrm{out}}=\AvgCons(\Pperp\Z{g}{\mathrm{in}},\graph,\Tcongd).$
    This allows us to use the same result of Proposition \ref{prop: ConsErr_prop} for $\Pperp\ConsErr{g}{\tau}$:
    \begin{align*}
        \|\Pperp\ConsErr{g}{\tau}\|_F\leqslant\econ\|\Pperp\InpErr{g}{\tau}\|_F.
    \end{align*}
    
    For the bound of  $\Pperp\InpErr{g}{\tau}$, we follow the similar procedure with the bound of  $\InpErr{g}{\tau}$. Using Eq.~\eqref{eqn: lem2 - InpErr},
    \begin{align*} 
        &\Pperp\InpErr{g}{\tau}=\frac{1}{L}\sum_{g'=1}^L(\Pperp\UconsErr{g,g'}{\tau-1})\\
        &\quad+\eta \Pperp\left(\sum_{g'\neq g}\Errgp{\tau}+\bbE[\sum_{g'\neq g}\Df{g'}{\tau}]\right)\\
        &\quad-\eta(L-1)\Pperp\left(\Errg{\tau}+\bbE[\Df{g}{\tau}]\right).
    \end{align*}
    Using part 4) of Proposition \ref{prop: Gradient deviation} and using the fact that both $\Pperp\boldsymbol{\Theta}^\star_{\setminus g}$ and $\Pperp\boldsymbol{\Theta}^\star_{g}$ are zero matrices,
    \begin{align*}
        &\Pperp\InpErr{g}{\tau}=\frac{1}{L}\sum_{g'=1}^L(\Pperp\UconsErr{g,g'}{\tau-1})\\
        &\quad +\eta \left(\Pperp\sum_{g'\neq g}\Errgp{\tau}
        +n\Pperp(\Ttasg{\tau}-\boldsymbol{\Theta}^\star_{\setminus g}){\B{\setminus g}{\tau}}^\top\right)\\
        &\quad
        -\eta(L-1)\left(\Pperp\Errg{\tau}+n\Pperp(\Ttag{\tau}-\boldsymbol{\Theta}^\star_g){\B{g}{\tau}}^\top)\right) \\  
        &=\frac{1}{L}\sum_{g'=1}^L(\Pperp\UconsErr{g,g'}{\tau-1})\\
            &\quad 
            +\eta \left(\Pperp\sum_{g'\neq g}\Errgp{\tau}
            +n\Pperp \Ttasg{\tau}{\B{\setminus g}{\tau}}^\top\right)\\
        &\quad
        -\eta(L-1)\left(\Pperp\Errg{\tau}+n\Pperp \Ttag{\tau}(\B{g}{\tau})^\top\right)\\
        &=\frac{1}{L}\sum_{g'=1}^L(\Pperp\UconsErr{g,g'}{\tau-1})\\
            &\quad 
            +\eta \left(\Pperp\sum_{g'\neq g}\Errgp{\tau}
            +n\Pperp \U{g}{\tau}\B{\setminus g}{\tau}{\B{\setminus g}{\tau}}^\top\right)\\
        &\quad
        -\eta(L-1)\left(\Pperp\Errg{\tau}+n\Pperp \Ttag{\tau}(\B{g}{\tau})^\top\right).
    \end{align*}
    
    Taking norm, using triangular and Cauchy-Schwarz inequalities, and using the fact that $\|\Pperp\|=1$,
    \begin{align*}
        &\|\Pperp\InpErr{g}{\tau}\|=\|\frac{1}{L}\sum_{g'=1}^L(\Pperp\UconsErr{g,g'}{\tau-1})\\
            &\quad +\eta \Pperp\sum_{g'\neq g}\Errgp{\tau}n\Pperp \Ttasg{\tau}{\B{\setminus g}{\tau}}^\top\\
            &\quad-\eta(L-1)\Pperp\left(\Errg{\tau}
            +m\Ttag{\tau}{\B{g}{\tau}}^\top\right)\|\\
        &\leqslant\|\frac{1}{L}\sum_{g'=1}^L(\Pperp\UconsErr{g,g'}{\tau-1})\|
            +\eta \|\Pperp\|\|\sum_{g'\neq g}\Errgp{\tau}\|\\
            &\quad+\eta n\|\Pperp \U{g}{\tau-1}\B{\setminus g}{\tau}{\B{\setminus g}{\tau}}^\top\|+\eta(L-1)\|\Pperp\|\|\Errg{\tau}\|
            \\
            &\quad+\eta n(L-1)\|\Pperp \U{g}{\tau-1}\B{g}{\tau}{\B{g}{\tau}}^\top\|\\
        &\leqslant\frac{1}{L}\cdot L\cdot \max_{g'}\|\Pperp\UconsErr{g,g'}{\tau-1}\|
            +\eta\|\sum_{g'\neq g}\Errgp{\tau}\|\\
            &\quad+\eta n\|\Pperp \U{g}{\tau-1}\|\|\B{\setminus g}{\tau}{\B{\setminus g}{\tau}}^\top\|
            +\eta(L-1)\|\Errg{\tau}\|\\
            &\quad+\eta n(L-1)\|\Pperp \U{g}{\tau-1}\|\|\B{g}{\tau}{\B{g}{\tau}}^\top\|.
        \end{align*}
        To bound the first term, we assume $\max_{g'}\|\Pperp\UconsErr{g,g'}{\tau-1}\|\leqslant\psig{\tau-1}$. For the second and fourth terms, we use parts 3) and 6) of Proposition \ref{prop: Gradient deviation}, and for the third and fifth terms, we use the definition of $\delg{\tau-1}$.
        \begin{align*}
        &\|\Pperp\InpErr{g}{\tau}\|\leqslant\psig{\tau-1}
            +\eta n(0.1\sigmin^2 +\maxsval{\B{\setminus g}{\tau}}^2)\delg{\tau-1}\\
            &\quad+\eta n(L-1)(0.1\sigmin^2 +\maxsval{\B{g}{\tau}}^2)\delg{\tau-1}\\
        &\leqslant\psig{\tau-1}
            +\eta n L\lr{ 0.1\sigmin^2 +\maxsval{\B{}{\tau}}^2}\delg{\tau-1}\\
        &\leqslant\psig{\tau-1}
            +\eta n L\lr{ 0.1\sigmax^2 +(1.1\sigmax)^2}\delg{\tau-1}\\
        &\leqslant\psig{\tau-1}
            +\frac{\ceta}{n\sigmax^2} nL\lr{1.31\sigmax^2}\delg{\tau-1}\\
        &\leqslant\psig{\tau-1}
            +1.4\ceta L\delg{\tau-1},
    \end{align*}
    where we used the claim that $\|\mat{M}'\|\leqslant\|\mat{M}\|$ if $\mat{M}'$ is $\mat{M}$ with some columns removed for the second inequality. For the third inequality, we used $\maxsval{\B{}{\tau}}\leqslant1.1\sigmax$ from part b) of Proposition \ref{prop: min-B}.
    Applying $\sqrt{r}$ for the Frobenius norm bound, 
    \begin{align*}
        \|\Pperp\InpErr{g}{\tau}\|_F\leqslant\sqrt{r}(\psig{\tau-1}+1.4\ceta L\delg{\tau-1}),    
    \end{align*}
     and thus {if $\psig{\tau-1}\leqslant \frac{0.1}{1.21\kappa^2}\delg{\tau-1}$}, then we have
    \begin{align}\label{eqn: PConsErr - Bound}
        &\|\Pperp\ConsErr{g}{\tau}\|_F\leqslant\econ\|\Pperp\InpErr{g}{\tau}\|_F\nonumber\\
        &\leqslant\econ\sqrt{r}(\psig{\tau-1}+1.4\ceta L\delg{\tau-1})\nonumber\\
        &\leqslant\econ\sqrt{r}(\frac{0.1}{1.21\kappa^2}+1.4\ceta L)\delg{\tau-1}\nonumber\\
        &=\econ\delg{\tau-1}\Cpcon,
    \end{align}
    where $\Cpcon:=\sqrt{r}(\frac{0.1}{1.21\kappa^2}+1.4\ceta L)$.
    This completes the proof. \qed

\subsection{Proof of Lemma \ref{lem: Bounding UconsErr}}\label{ssc: proof of lemma 3}
    Lemma \ref{lem: Bounding UconsErr} provides bounds for $\|\UconsErr{g,g'}{\tau}\|_F=\|\U{g}{\tau}-\U{g'}{\tau}\|_F$ and $\|\Pperp\UconsErr{g,g'}{\tau}\|_F=\|\Pperp(\U{g}{\tau}-\U{g'}{\tau})\|_F$.  We continue to assume that $\Tcongd$ satisfies Eq.~\eqref{eqn: Tcongd - min}, and use the perturbed QR factorization (Proposition \ref{prop: Perteurbed QR}).

   \subsubsection{Bounding $\|\UconsErr{g,g'}{\tau}\|_F$}
    Consider $\Utilde{g}{\tau}-\Utilde{g'}{\tau}$.
    By adding and subtracting $\Uhat{\tau}$,
    \begin{align*}
        \Utilde{g}{\tau}-\Utilde{g'}{\tau}
        &=\Utilde{g}{\tau}-\Uhat{\tau}-\Utilde{g'}{\tau}+\Uhat{\tau}\\
        &=\ConsErr{g}{\tau}-\ConsErr{g'}{\tau}.
    \end{align*}
    Apply Frobenius norm, then by triangular inequality and Lemma \ref{lem: Conserr} (Eq.~\eqref{eqn: ConsErr - Bound}),
    \begin{align}
        \|\Utilde{g}{\tau}-\Utilde{g'}{\tau}\|_F
        &\leqslant\|\ConsErr{g}{\tau}\|_F+\|\ConsErr{g'}{\tau}\|_F\nn\\
        &\leqslant2\econ\Ccon\label{eqn: lem3 - uutilde}.
    \end{align}
        
    Apply Proposition \ref{prop: Perteurbed QR} (Eq.~\eqref{eqn: Perturbed QR-Q}) with $\mat{Z}_1\equiv\U{g}{\tau}$, $\mat{Z}_2\equiv\U{g'}{\tau}$, $\tilde{\mat{Z}}_1\equiv \Utilde{g}{\tau}$, and $\tilde{\mat{Z}}_2\equiv\Utilde{g'}{\tau}$. 
    Under the conditions $\ceta=0.4$,
    $\delg{\tau-1}\leqslant0.02/\sqrt{r}\kappa^2$, $\rhog{\tau-1}\leqslant0.1\ceta/\kappa^2$,
    and $\|\ConsErr{g}{\tau}\|_F\leqslant\ccon$
    (needed to use Eq.~\eqref{eqn: SD-den} for lower bounding $\minsval{\Utilde{g}{\tau}}$), 
    we have
    \begin{align}%\label{eqn: UconsErr}
        &\|\UconsErr{g,g'}{\tau}\|_F\leqslant
            2\sqrt{2}\frac{\|\Utilde{g}{\tau}-\Utilde{g'}{\tau}\|_F}{\minsval{\Utilde{g'}{\tau}}}\leqslant 
            2\sqrt{2}\frac{\econ\Ccon}
        {1-0.15\ceta/\kappa^2}\nonumber \\
        &\leqslant 
            2\sqrt{2}\frac{\econ\Ccon}
        {1-0.15\cdot 0.4/\kappa^2}
        \leqslant 3.1\econ\Ccon=\econ\Cucon, \nonumber
    \end{align}
    where $\Cucon:=3.1\Ccon=\sqrt{r}\frac{0.31\ceta+0.11\ceta L}{\kappa^2}$.

    \subsubsection{Bounding $\|\Pperp\UconsErr{g,g'}{\tau}\|_F$} 
    Using the QR relations $\Utilde{g}{\tau}\qreq\U{g}{\tau}\R{g}{\tau}$ and $\Utilde{g'}{\tau}\qreq\U{g'}{\tau}\R{g'}{\tau}$, 
    we have
    \begin{align*}
        &\Pperp(\U{g}{\tau}-\U{g'}{\tau})=\Pperp\left(\Utilde{g}{\tau}{(\R{g}{\tau})}^{-1}
            -\Utilde{g'}{\tau}{(\R{g'}{\tau})}^{-1}\right).
    \end{align*}
    Adding and subtracting $\Pperp\Utilde{g'}{\tau}(\R{g}{\tau})^{-1}$ and $\Pperp\Uhat{\tau}(\R{g}{\tau})^{-1}$ and using the fact, for invertible matrices $\mat{A}$ and $\mat{B}$, $\mat{A}^{-1}-\mat{B}^{-1}=\mat{A}^{-1}\mat{B}\mat{B}^{-1}-\mat{A}^{-1}\mat{A}\mat{B}^{-1}=\mat{A}^{-1}(\mat{B}-\mat{A})\mat{B}^{-1}$, 
    %we have
    \begin{align*}
        &\Pperp(\U{g}{\tau}-\U{g'}{\tau})=\Pperp(\Utilde{g}{\tau}-\Utilde{g'}{\tau}){(\R{g}{\tau})}^{-1}\\
        &\quad +\Pperp\Utilde{g'}{\tau}({(\R{g}{\tau})}^{-1}-{(\R{g'}{\tau})}^{-1})\\
        &=\Pperp\left(\Utilde{g}{\tau}-\Uhat{\tau}+\Uhat{\tau}-\Utilde{g'}{\tau}\right){(\R{g}{\tau})}^{-1}\\
            &\quad 
            +\Pperp\Utilde{g'}{\tau}\left({(\R{g}{\tau})}^{-1}-{(\R{g'}{\tau})}^{-1}\right)\\
        &=\Pperp(\ConsErr{g}{\tau}-\ConsErr{g'}{\tau}){(\R{g}{\tau})}^{-1}\\
            &\quad 
            +\Pperp\Utilde{g'}{\tau}\left({(\R{g}{\tau})}^{-1}({\R{g'}{\tau}}-{\R{g}{\tau}}){(\R{g'}{\tau})}^{-1}\right).
    \end{align*}
    By taking Frobenius norm and using triangular inequality,
    \begin{align}
        &\|\Pperp(\U{g}{\tau}-\U{g'}{\tau})\|_F\nonumber\\
        &\leqslant\|\Pperp(\ConsErr{g}{\tau}-\ConsErr{g'}{\tau}){(\R{g}{\tau})}^{-1}\|_F\nonumber\\
            &\quad +\|\Pperp\Utilde{g'}{\tau}{(\R{g}{\tau})}^{-1}({\R{g'}{\tau}}-{\R{g}{\tau}}){(\R{g'}{\tau})}^{-1}\|_F\nonumber\\
        &\leqslant 2\frac{\|\Pperp\ConsErr{g}{\tau}\|_F}{\minsval{\Utilde{g}{\tau}}}\nonumber\\
            &\quad
            +\|\Pperp\Utilde{g'}{\tau}\| \|{(\R{g}{\tau})}^{-1}\| \|{\R{g'}{\tau}}-{\R{g}{\tau}}\|_F \|{(\R{g'}{\tau})}^{-1}\|\nonumber\\
             % \end{align}
             %  \begin{align}
        &\leqslant 2\frac{\|\Pperp\ConsErr{g}{\tau}\|_F}{\minsval{\Utilde{g}{\tau}}}
            +\frac{\|\Pperp\Utilde{g'}{\tau}\|\|{\R{g'}{\tau}}-{\R{g}{\tau}}\|_F}{\minsval{\R{g}{\tau}}\minsval{\R{g'}{\tau}}}\nonumber\\
        &\leqslant 2\frac{\|\Pperp\ConsErr{g}{\tau}\|_F}{\minsval{\Utilde{g}{\tau}}}
            +\frac{\|\Pperp\Utilde{g'}{\tau}\|\|{\R{g'}{\tau}}-{\R{g}{\tau}}\|_F}{\minsval{\Utilde{g}{\tau}}\minsval{\Utilde{g'}{\tau}}}\nn
    \end{align}
    where $\|(\R{g}{\tau})^{-1}\|=\frac{1}{\minsval{\Utilde{g}{\tau}}}$. 
    Using known bounds for $\|\Pperp\Utilde{g}{\tau}\|\leqslant(1-0.6\ceta/\kappa^2)\delg{\tau-1}\leqslant\delg{\tau-1}$ from Eq.~\eqref{eqn: SD-num result}, $\minsval{\Utilde{g}{\tau}}\geqslant1-0.15\ceta/\kappa^2\geqslant0.94$ with $\ceta=0.4$ from Eq.~\eqref{eqn: SD-den result}, and $\|\Pperp\ConsErr{g}{\tau}\|_F\leqslant\econ\Cpcon\delg{\tau-1}$ from  Eq.~\eqref{eqn: PConsErr - Bound}, we obtain
    \begin{align}
        &\|\Pperp(\U{g}{\tau}-\U{g'}{\tau})\|_F\nn\\
        &\quad \leqslant (\frac{2\econ \Cpcon}{0.94}+\frac{\|{\R{g'}{\tau}}-{\R{g}{\tau}}\|_F}{0.94^2})\delg{\tau-1}.\label{eq:perturbed qr 2 - result}
    \end{align}

    For the bound of $\|{\R{g'}{\tau}}-{\R{g}{\tau}}\|_F$, we apply Proposition \ref{prop: Perteurbed QR} (Eq.~\eqref{eqn: Perturbed QR-R}) with $\mat{R}_1\equiv\R{g}{\tau}$, $\mat{R}_2\equiv\R{g'}{\tau}$, $\tilde{\mat{Z}}_1\equiv \Utilde{g}{\tau}$, and $\tilde{\mat{Z}}_2\equiv\Utilde{g'}{\tau}$.
    Using $\|\mat{P}_{\Utilde{g}{\tau}}\| = \|\mat{P}_{\U{g}{\tau}}\|=1 $ since $\textrm{col}(\Utilde{g}{\tau})=\textrm{col}(\U{g}{\tau})$ and $\U{g}{\tau}$ is orthonormal,
    \begin{align*}
        &\|{\R{g'}{\tau}}-{\R{g}{\tau}}\|_F
        \leqslant\sqrt{2}\|\R{g}{\tau}\|\kappa_2(\Utilde{g}{\tau})
            \frac{\|\mat{P}_{\Utilde{g}{\tau}}(\Utilde{g'}{\tau}-\Utilde{g}{\tau})\|_F}{\maxsval{\Utilde{g}{\tau}}}\nonumber\\
        &\leqslant\sqrt{2}\|\R{g}{\tau}\|\frac{\maxsval{\Utilde{g}{\tau}}}{\minsval{\Utilde{g}{\tau}}}
            \frac{\|\mat{P}_{\Utilde{g}{\tau}}\|\|\Utilde{g'}{\tau}-\Utilde{g}{\tau}\|_F}{\maxsval{\Utilde{g}{\tau}}}\nonumber\\
        &\leqslant\sqrt{2}\|\R{g}{\tau}\|
        \frac{\|\mat{P}_{\U{g}{\tau}}\|\|\Utilde{g'}{\tau}-\Utilde{g}{\tau}\|_F}{\minsval{\Utilde{g}{\tau}}}\nonumber\\
        &\leqslant2\sqrt{2}\maxsval{\Utilde{g}{\tau}}\frac{\|\ConsErr{g}{\tau}\|_F}{\minsval{\Utilde{g}{\tau}}}\quad(\textrm{by~Eq.}~\eqref{eqn: lem3 - uutilde}).\nonumber
    \end{align*}
    {The upper bound of $\maxsval{\Utilde{g}{\tau}}$ can be obtained by
    \begin{align*}
        &\maxsval{\Utilde{g}{\tau}}=\|\Utilde{g}{\tau}-\Uhat{\tau}+\Uhat{\tau}\|=\|\Uhat{\tau}+\ConsErr{g}{\tau}\|\\
        &\leqslant\|\Uhat{\tau}\|+\|\ConsErr{g}{\tau}\|_F\\
        &\leqslant \|\Ubar{\tau-1}-\eta\gradU{\tau}\|+\|\ConsErr{g}{\tau}\|_F\\
        &\leqslant \|\Ubar{\tau-1}\|+\eta(\|\Err{\tau}\|+\|\bbE[\gradU{\tau}]\|)+\|\ConsErr{g}{\tau}\|_F.
    \end{align*}
    Set $\ceta=0.4$ and assume that $\|\ConsErr{g}{\tau}\|_F\leqslant\ccon$ and $\delg{\tau-1}\leqslant \frac{0.02}{\sqrt{r}\kappa^2}$. 
    Using parts 2) and 3) of Proposition \ref{prop: Gradient deviation} for $\|\Err{\tau}\|$ and $\|\mathbb{E}[\gradU{\tau}]\|$ with $\epsilon_1=0.1$, 
    \begin{align*}
        &\maxsval{\Utilde{g}{\tau}}
        \leqslant1+\eta1.64 n\delg{\tau-1}\sigmax^2+\ccon\\
        &\leqslant 1+1.64\frac{\ceta}{n\sigmax^2}n\sigmax^2\frac{0.02}{\sqrt{r}\kappa^2}+\frac{0.01\ceta}{\kappa^2}\\
        &\leqslant 1+0.0328\ceta/\kappa^2+\frac{0.01\ceta}{\kappa^2}\leqslant 1+0.05\ceta/\kappa^2\leqslant 1.1
    \end{align*}
    where the first inequality used $\|\U{g}{\tau-1}\|=1$ and $\sigmax\geqslant\sigmin$, and the third inequality used $\sqrt{r}>1$.} 
    Combined with Eq.~\eqref{eqn: ConsErr - Bound}, 
    \begin{align*}
        &\|{\R{g'}{\tau}}-{\R{g}{\tau}}\|_F\leqslant2\sqrt{2}(1.1)\frac{\econ\Ccon}{0.94}.
    \end{align*}
    Substituting the above into Eq.~\eqref{eq:perturbed qr 2 - result} yields
    \begin{align*}
        &\|\Pperp(\U{g}{\tau}-\U{g'}{\tau})\|_F\\
        &\leqslant\econ\delg{\tau-1}\lr{\frac{2\Cpcon}{0.94}
            +\frac{2\sqrt{2}(1.1)\Ccon}{0.94^3}}\\
        &=\econ\delg{\tau-1}(2.2\Cpcon+3.8\Ccon)
        \leqslant\econ\delg{\tau-1}\Cpucon
    \end{align*}
    where $2.2\Cpcon+3.8\Ccon\leqslant\sqrt{r}(1+5.5\ceta L)=:\Cpucon$.
    This completes the proof. 
    \qed

    \begin{figure*}[ht]
        % \centering
        \begin{subfigure}{0.35\textwidth}
            \centering
            \includegraphics[width=\linewidth]{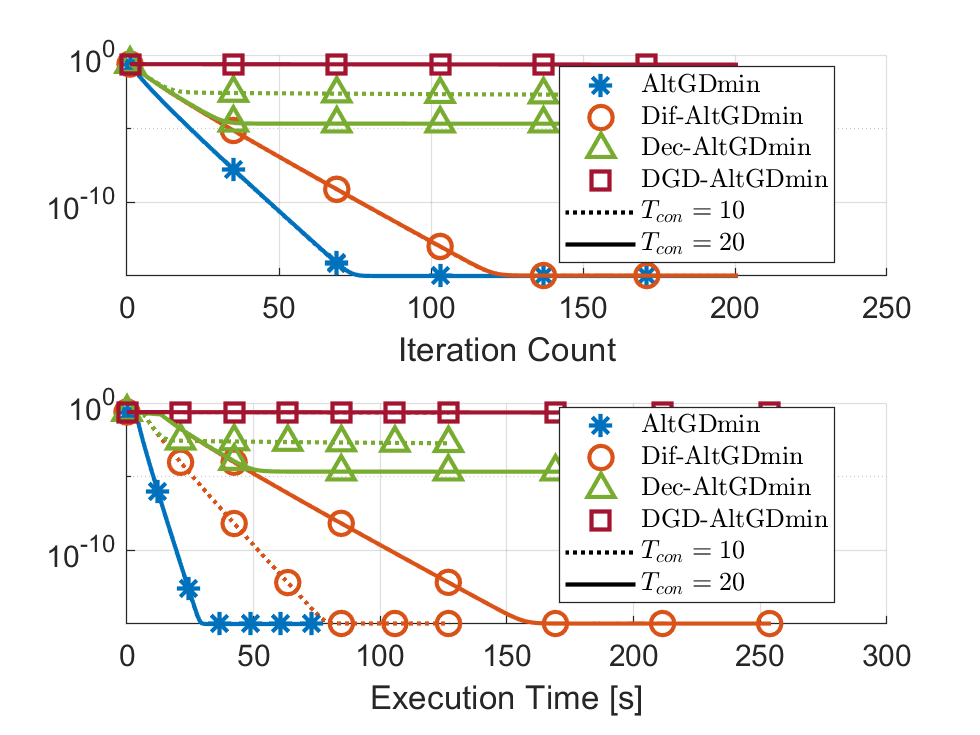}
            \vspace{-6 mm}
            \caption{$\Tcon=10$ and $20$}
            \label{fig:1a}
        \end{subfigure}
        \hspace{-1 mm}
       % \hfill
        \begin{subfigure}{0.35\textwidth}
            \centering
            \includegraphics[width=\linewidth]{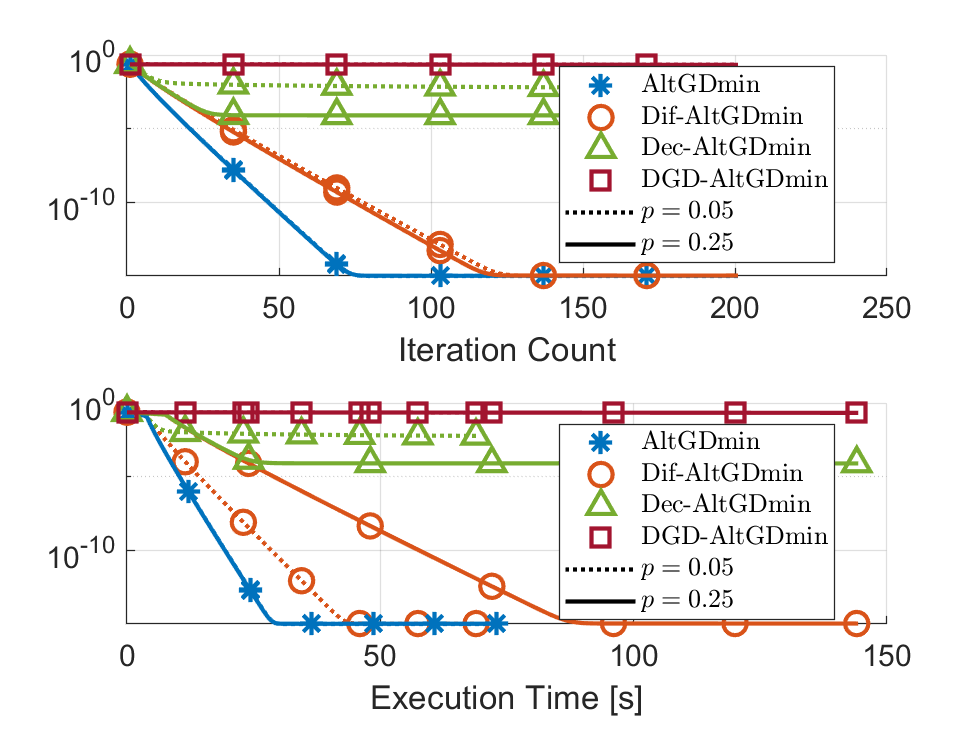}
            \vspace{-6 mm}
            \caption{$p=0.05$ and $0.25$}
            \label{fig:1b}
        \end{subfigure}
         \hspace{-6 mm}
       % \hfill
        \begin{subfigure}{0.35\textwidth}
            \centering
            \includegraphics[width=\linewidth]{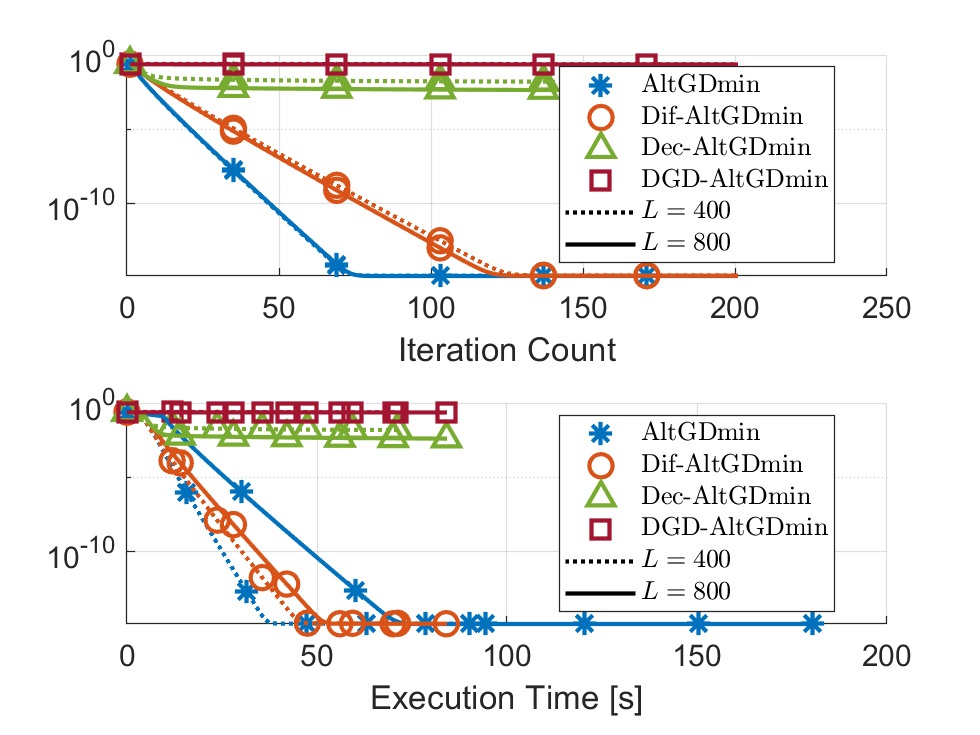}
            \vspace{-6 mm}
            \caption{$L=400$ and $800$}
            \label{fig:1c}
        \end{subfigure}
        \vspace{-5 mm}
        \caption{Subspace distance vs. iteration count and execution time in seconds. In all plots, y-axis is $\SD{\U{1}{\tau}}{\Ustar}$. We compare the performance of algorithms by under different communication settings -- $\Tcon:=\Tconinit=\Tcongd$, $p$, and $L$. The default setting is $\Tcon=5$, $\Tgd=200$, $L=300$, $d=300,\ T=800,\ r=4,\ n=50$, and $p=0.03$, and we vary one parameter at a time while fixing all other parameters at their default values.
        }\label{fig:Varying Tcon}
    \end{figure*}
\section{Simulations}\label{sec: simulations}
\begin{figure*}[ht]
    % \centering
    \begin{subfigure}{0.35\textwidth}
        \centering
        \includegraphics[width=\linewidth]{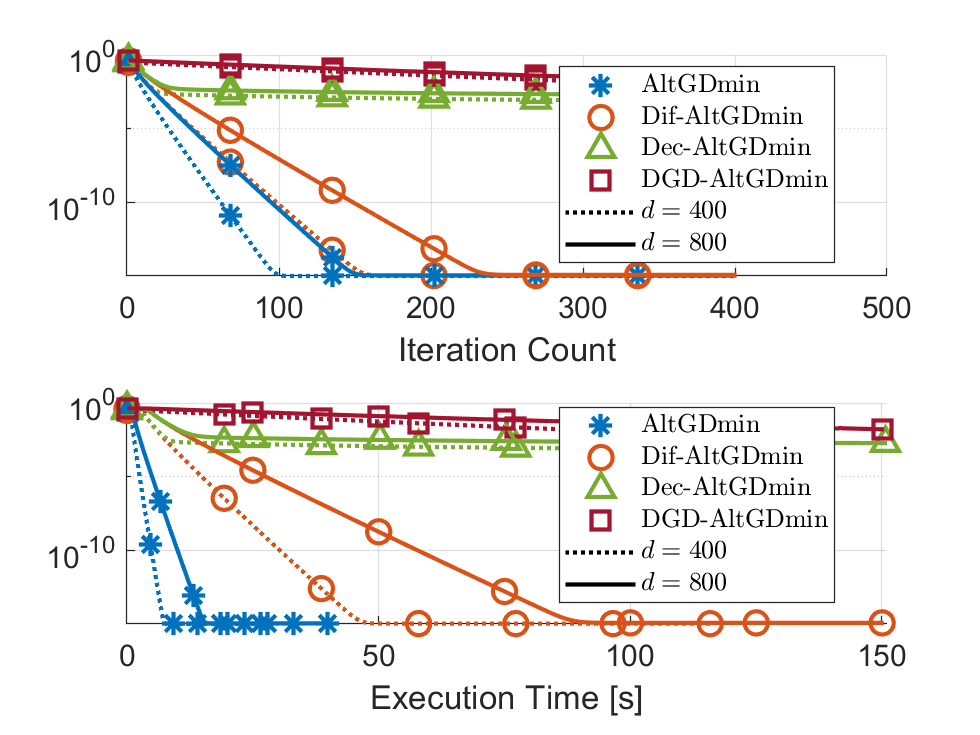}
        \vspace{-6 mm}
        \caption{$d=400$ and $800$}
        \label{fig:2a}
    \end{subfigure}
        \hspace{-1 mm}
   % \hfill
    \begin{subfigure}{0.35\textwidth}
        \centering
        \includegraphics[width=\linewidth]{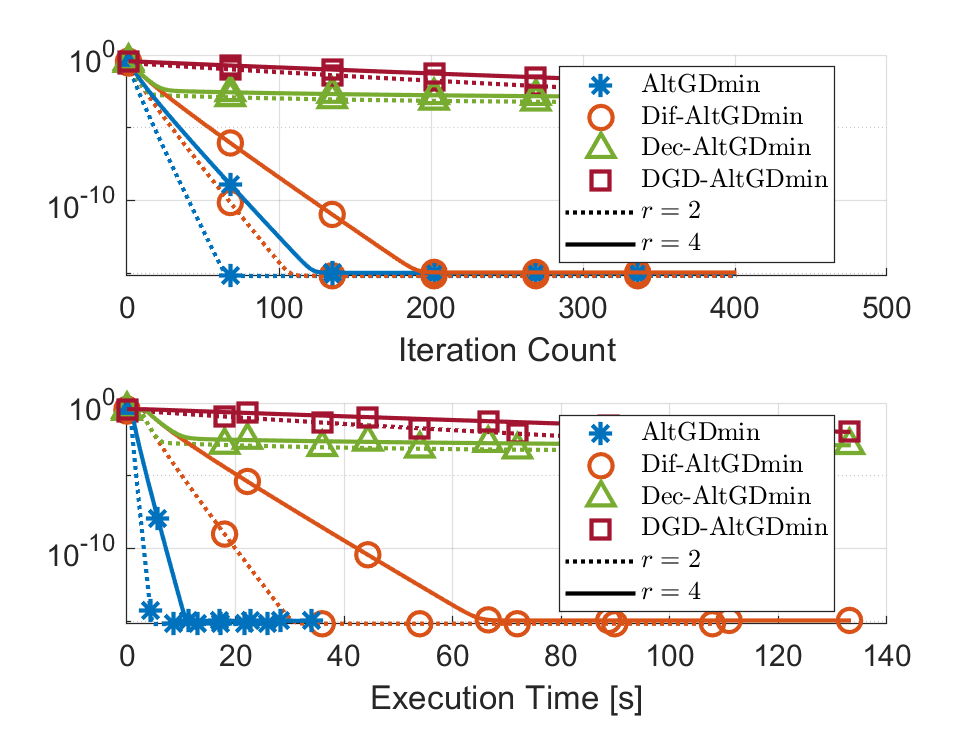}
        \vspace{-6 mm}
        \caption{$r=2$ and $4$}
        \label{fig:2b}
    \end{subfigure}
        \hspace{-6 mm}
   % \hfill
    \begin{subfigure}{0.35\textwidth}
        \centering
        \includegraphics[width=\linewidth]{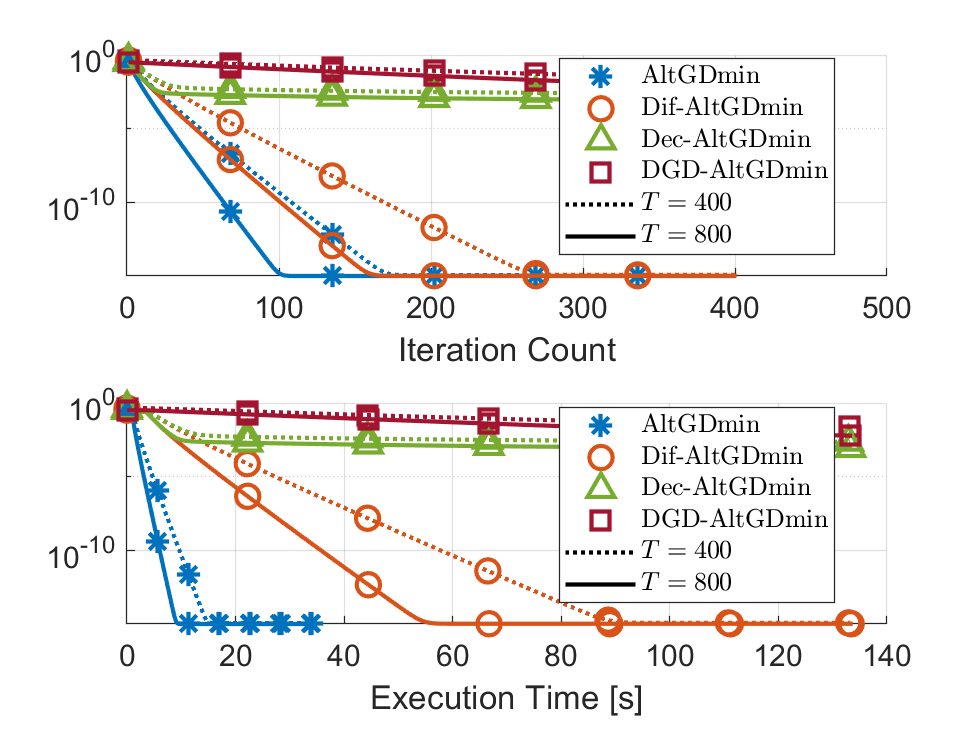}
        \vspace{-6 mm}
        \caption{$T=400$ and $800$}
        \label{fig:2c} 
    \end{subfigure}
    \vspace{-6 mm}
    \caption{Subspace distance vs. iteration count and execution time. In all plots, y-axis is $\SD{\U{1}{\tau}}{\Ustar}$. We compare the impact of problem parameter sizes-- $d,\ r,$ and $T$. The default setting is $\Tcongd=\Tconinit=5, \Tgd=400$, $L=20$, $d=T=600, r=4$, and  $n=50$, and we vary one parameter at a time while fixing all other parameters at their default values.}
    \label{fig: Experiment 2}
\end{figure*}

\noindent{\bf Data and network settings.}
All simulations are implemented in MATLAB using synthetic data. 
The communication network $\graph$ and the dataset $\{\Xt,\yt\}_{t=1}^T$ are randomly generated.
Although our theoretical analysis relies on sample-splitting, it is not enforced in the simulations.
The network is modeled as an Erdős-Rényi graph with $L$ nodes and connection probability $p$.
To emulate the communication overheads, we adopt a standard latency-bandwidth model in which the communication time scales linearly with the message size and the local node degree. 
Assuming a network bandwidth of $150\, \rm Mbps$ and a latency of $20\, m \rm s$, the communication time per agreement round is approximated as $t_{\rm comm}=20\times 10^{-3} + \frac{8dr\cdot\max_g\degg}{150\times 10^6}\;\;\rm seconds$.
 We used same agreement rounds for initialization and GD steps, i.e.,  $\Tcon:=\Tconinit=\Tcongd$ 
 The step size is set to $\eta=\frac{0.4}{n\hat{\sigma}_{\rm max}^{\star^2}}$, where $\hat{\sigma}_{\rm max}^{\star^2}$ is the largest diagonal entry of $\R{g}{\Tpm}$.
For the evaluation, we report the empirical average of the subspace distances $\SD{\U{1}{\tau}}{\Ustar}$ over 100 independent trials, plotted against either iteration count or execution time.
We compared the performance of our Dif-AltGDmin algorithm with three baselines: (i) AltGDmin \cite{nayer2022fast},  a centralized algorithm that aggregates gradients from all $L$ nodes in one communication step and then broadcasts the updated $\mat{U}$, (ii) Dec-AltGDmin \cite{moothedath2022fast}, and {(iii) a DGD-variation of AltGDmin, defined as $\Utilde{g}{\tau}\leftarrow QR(\frac{1}{\degg}\sum_{g'\in\Ng} \U{g'}{\tau-1}-\eta\Df{g}{\tau})$.}
For centralized method, the communication cost scales linearly with total number of agents $L$ rather than $\max_g\degg$.
{Based on this setting, we conduct two sets of experiments. Experiment 1, investigates the network-related parameters ($\Tcon,\ p,\ L$). Experiment 2 examines the impact of data dimensions ($d,\ r,\ T$).}

\noindent{\bf Experiment 1.} 
In this experiment, we studied the impact of network and communication parameters -- $\Tcon$, $p$, and $L$ --  on convergence and communication efficiency.
{ Unless otherwise specified, the parameters are set to $L=300$, $d=300$, $T=800$, $n=50$, and $r=4$.}
As illustrated in Figs.~\ref{fig:1a}-\ref{fig:1c}, the proposed Dif-AltGDmin converges to the same accuracy level as AltGDmin. 
The Dif-AltGDmin attains the final error level even for smaller values of $\Tcon$ or $p$, whereas
Dec-AltGDmin stagnates above bounds that strongly depend on these parameters.
Specifically, its performance improves with larger number of agreement rounds or denser connectivity, yet remains significantly less accurate than that of AltGDmin and Dif-AltGDmin.
This results highlight the robustness of Dif-AltGDmin to limited communication and weak network connectivity as discussed in Remarks \ref{remark: Tcon} and \ref{remark: robustness to connectivity}.
The DGD-variant fails to converge effectively, consistent with the observation reported in \cite{moothedath2022fast}.
Fig.~\ref{fig:1c} shows that, although AltGDmin requires fewer iterations than Dif-AltGDmin to converge, its execution time increases significantly with network size due to its linear communication cost in $L$.
In contrast, the per-node execution time of Dif-AltGDmin scales with local degree $\degg$, making it more favorable for large networks. 
This result demonstrates that Dif-AltGDmin outperforms the centralized approach as network size increases and communication scales, since the centralized server increasingly becomes a bottleneck when interacting with many nodes.

\noindent{\bf Experiment 2.} 
In this experiment, we investigate the impact of problem parameter sizes on the convergence of the proposed algorithm.
The default setting is $\Tgd=400$, $L=20$,\ $d=T=600,\ r=4$ and  $n=50$.
Recall that the optimization variables are $\mat{U}\in\mathbb{R}^{d\times r}$ and $\mat{b}_t \in\mathbb{R}^{r}, \ t\in[T]$. As the problem dimension increases, the estimation and communication process become more challenging.
As can be seen in Figs.~\ref {fig:2a} and \ref{fig:2b}, increasing either $d$ or $r$ results in an increase in both the number of iterations and the execution time for all algorithms.
Next, we examine the effect of varying the number of tasks $T$ in Fig.~\ref{fig:2c}.
Increasing the number of tasks leads to faster convergence in terms of both iterations and execution time. 
With the per task sample size $n$ fixed, increasing the number of tasks $T$ increases the total available data, thereby enabling more effective collaboration.
As a result, the algorithms achieve accuracy level with fewer iterations. 
Moreover, as discussed in Section~\ref{ssc: complexities}, the execution time depends on the degree of the node rather on $T$. 
Consequently, increasing the number of tasks does not increase the per-iteration communication cost.
In all cases, Dif-AltGDmin achieves performance comparable to that of the centralized AltGDmin.

\section{Conclusion}\label{sec: conclusion}
We proposed \emph{Dif-AltGDmin}, a diffusion-based decentralized multi-task representation learning algorithm that combines local GD and minimization with efficient decentralized information aggregation. We established convergence guarantees and characterized the sample complexity of the proposed method. Furthermore, we analyzed the time and communication cost and identified regimes in which the decentralized approach outperforms centralized AltGDmin approach.
We conducted simulations by varying the problem dimensions and network parameters, and evaluated the performance of the proposed approach against decentralized baselines and a centralized method.
Future work will focus on further reducing communication overhead by integrating techniques such as quantization, compression, and sporadic communication. 

\vspace{-2 mm}
\section*{References}
\vspace{-2 mm}
\bibliographystyle{IEEEtran}
\bibliography{references.bib, bandits.bib}
\begin{wrapfigure}{l}{25mm} 
\includegraphics[width=1in,height=1.25in,clip,keepaspectratio]{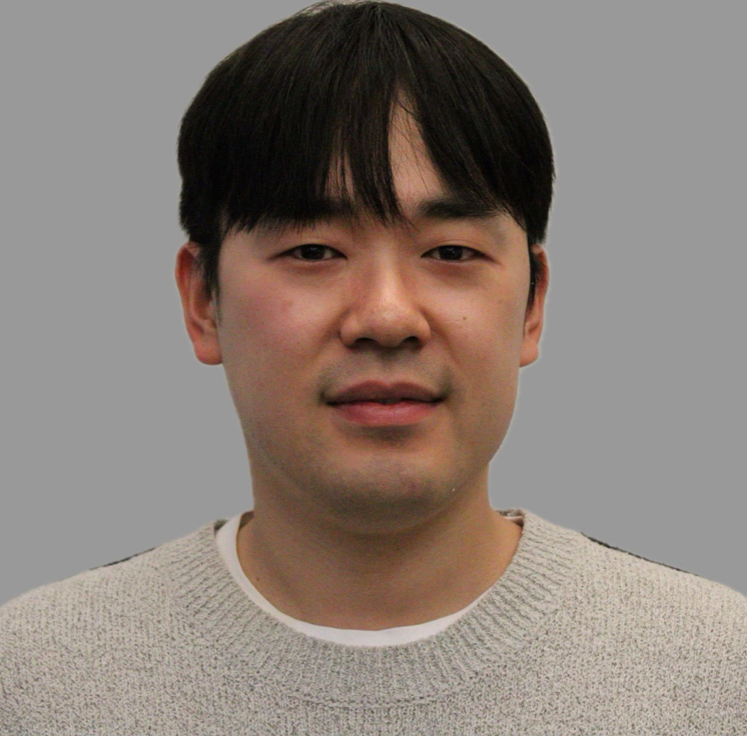}
\end{wrapfigure}\par
\textbf{Donghwa Kang} received the B.S. degree in Mechanical and Control Engineering from Handong Global University, Pohang, South Korea, in 2021, and the M.S. degree in Mechanical Engineering from Ulsan National Institute of Science and Technology, Ulsan, South Korea, in 2023. He is currently a Ph.D student in the Department of Electrical and Computer Engineering at Iowa State University, Ames, IA, USA. His research interests include decentralized multi-task learning, sparse signal processing, and reinforcement learning.

\begin{wrapfigure}{l}{25mm} 
\includegraphics[width=1in,height=1.25in,clip,keepaspectratio]{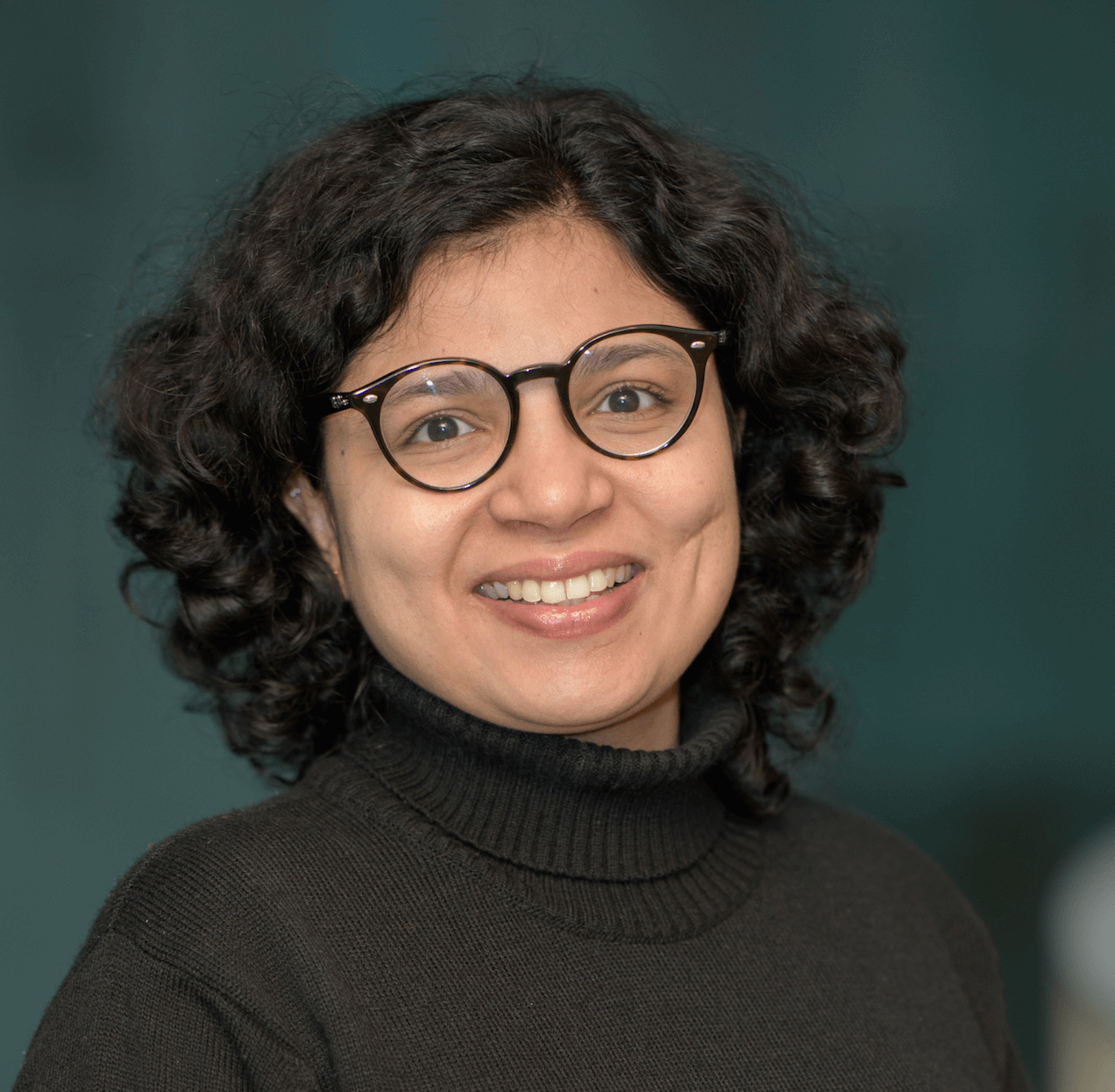}
\end{wrapfigure}\par
\textbf{Shana Moothedath} (Senior Member, IEEE) received her Ph.D. degree in Electrical Engineering from the Indian Institute of Technology Bombay (IITB) in 2018, and she was a postdoctoral scholar in Electrical and Computer Engineering at the University of Washington, Seattle till 2021. Currently, she is Harpole-Pentair Assistant Professor of Electrical and Computer Engineering at Iowa State University. Her research focuses on distributed decision-making, control and security of dynamical systems, and  signal processing. She received the NSF CAREER Award in 2025, the Best Research Thesis Award at IITB in 2019, and selected as a MIT-EECS Rising Star in 2019.

\end{document}